\documentclass[final,5p,times,twocolumn]{elsarticle}

\usepackage{amssymb}
\usepackage{amsthm}

\usepackage{longtable}
\usepackage{subcaption}
\usepackage{hyperref}
\usepackage{amsmath}

\newcommand{\h}{\mathbf{h}}
\newcommand{\x}{\mathbf{x}}
\newcommand{\W}{\mathbf{W}}
\newcommand{\D}{\mathbf{D}}
\newcommand{\Wh}{\mathbf{W}_h}
\newcommand{\Wx}{\mathbf{W}_x}
\newcommand{\I}{\mathbf{I}}
\newcommand{\J}{\mathbf{J}}
\renewcommand{\b}{\mathbf{b}}

\newcommand{\R}{\mathbb{R}}

\newtheorem{corollary}{Corollary}

\newtheorem{proposition}{Proposition}

\journal{Journal}

\begin{document}

\begin{frontmatter}


\author{Claudio Gallicchio\corref{cor1}}
\ead{gallicch@di.unipi.it}
\address{University of Pisa, Department of Computer Science\\
Largo Bruno Pontecorvo, 3, Pisa, 56127, Italy}

\title{Euler State Networks: Non-dissipative Reservoir Computing}

\begin{abstract}
Inspired by the numerical solution of ordinary differential equations,
in this paper we propose a novel Reservoir Computing (RC) model, called the Euler State Network (EuSN). The presented approach makes use of forward Euler discretization and antisymmetric recurrent matrices to design reservoir dynamics that are both stable and non-dissipative by construction.

Our mathematical analysis shows that the resulting model is biased towards a unitary effective spectral radius and zero local Lyapunov exponents, intrinsically operating near to the edge of stability. Experiments on long-term memory tasks show the clear superiority of the proposed approach over standard RC models in problems requiring effective propagation of input information over multiple time-steps.
Furthermore, results on time-series classification benchmarks indicate that EuSN is able to match (or even exceed) the accuracy of trainable Recurrent Neural Networks, while retaining the training efficiency of the RC family, resulting in up to $\approx 490$-fold savings in computation time and $\approx 1750$-fold savings in energy consumption.

\end{abstract}


\end{frontmatter}


\section{Introduction}
\label{sec.introduction}

The interest in studying neural network architectures from a dynamical system perspective has recently been attracting increasing research attention \cite{chen2018neural,chang2018reversible,chang2017multi,lu2018beyond}. 
The key insight is that the computation performed by some kinds of neural networks, e.g., residual networks, can be understood as the numerical solution to an ordinary differential equation (ODE) through discretization \cite{haber2017stable}. 
This intuitively simple observation brings about the possibility of imposing desirable properties in the behavior of the neural network by imposing specific conditions to the corresponding ODE. Stability plays a key role in this sense, being related to the propagation of both input signals, during inference, and gradients, during training. 

In this paper, we focus on Recurrent Neural Network (RNN) architectures, and especially on Reservoir Computing (RC) \cite{Verstraeten2007, Lukosevicius2009}. 
RC establishes a particularly attractive approach for the design and training of RNNs, where the hidden recurrent \emph{reservoir} layer is left untrained, leaving the entire training effort to the output layer alone. 
The stability of forward signal propagation through the reservoir is therefore of great importance. 
Due to the fact that the parameters 
that determine the dynamics of the reservoir
are untrained, some form of constraint is indeed necessary to avoid instability when the network is operated with a driving input. This aspect is related to the widely known 
fading memory property of RC networks, 
which makes it difficult to maintain the input signal information in the state dynamics across several time-steps.
%
Despite this limitation, RC has become more and more popular because of the impressively advantageous trade-off between predictive performance and training efficiency. As a result,
it is often the chosen approach for embedded applications distributed at the edge \cite{dragone2015cognitive,bacciu2014experimental}, as well as for RNN implementations in neuromorphic hardware \cite{wang2023echo, tanaka2019recent, torrejon2017neuromorphic, markovic2020quantum}.
However, the performance gap with fully trainable state-of-the-art RNNs still leaves space for improvements.

This paper extends our previous work in \cite{gallicchio2021esann},
introducing and analyzing a new RC method that is inspired by the numerical solution of ODEs.
As the reservoir architecture is obtained by forward Euler discretization of an intrinsically stable ODE, the resulting model is called the Euler State Network (EuSN).
The initialization condition imposed to the reservoir inherently yields state dynamics that are biased to be both stable and non-dissipative, thereby mitigating the lossy transmission of input signals over time, sensibly reducing the performance gap with fully trainable RNNs in time-series classification tasks.

The contributions of this work can be summarized as follows:
\begin{itemize}
\item We introduce the EuSN model, a novel approach to design RC neural network architectures based on ODE discretization. Overcoming the natural limitations of standard RC models, EuSN is specifically designed to fulfill both stability and non-dissipative properties.
\item We derive a mathematical characterization of the proposed reservoir system, focusing on its asymptotic stability and local Lyapunov exponents. 
\item We provide an extensive experimental evaluation of EuSN. First, we shed light on its long-term memorization abilities in comparison to standard RC models. Then, we assess the accuracy-efficiency trade-off of EuSN in comparison to both RC and fully trainable RNN models on several time-series classification benchmarks.
\end{itemize}

The rest of this paper is organized as follows. In Section~\ref{sec.esn} we give an overview on the basic concepts of RC.
We introduce the EuSN model 
in Section~\ref{sec.model}, and we elaborate its mathematical analysis in  Section~\ref{sec.analysis}. The experimental assessment of the approach is provided in Section~\ref{sec.experiments}.
Finally, Section~\ref{sec.conclusions} concludes the paper.


\section{Reservoir Computing and Echo State Networks}
\label{sec.esn}
Reservoir Computing (RC) is a common denomination for a class of recurrent neural models that are based on untrained internal dynamics \cite{Verstraeten2007,Lukosevicius2009,tanaka2019recent}. In general, the architecture comprises a fixed hidden recurrent layer called \emph{reservoir}, and an output \emph{readout} layer that is the only trainable component. Here, in particular, we focus on the Echo State Network (ESN) model \cite{jaeger2004harnessing, Jaeger2001}, where the reservoir operates in discrete time-steps and typically makes use of $\tanh$ non-linearity. We consider a reservoir architecture with $N$ reservoir neurons and $X$ input units, using $\h(t)$ and $\x(t)$ to denote, respectively, the state of the reservoir and the external input at time-step $t$. 
Referring to the general ESN formulation with leaky integrator neurons  \cite{jaeger2007optimization}, the reservoir state is evolved according to the following iterated map:
\begin{equation}
\label{eq.esn}
\h(t) = (1-\alpha) \, \h(t-1) +
\alpha \, \tanh(\Wh\;\h(t) + \Wx\;\x(t)+\mathbf{b}),
\end{equation}
where $\alpha \in (0,1]$ is the leaking rate, $\Wh \in \R^{N \times N}$ is the recurrent reservoir weight matrix, $\Wx \in \R^{N \times X}$ is the input weight matrix, and $\mathbf{b} \in \R^N$ is a vector of bias weights.
As an initial condition, the reservoir state is typically started at the origin, i.e., the null vector: $\h(0) = \mathbf{0}$.

Interestingly, all the weight values that parametrize the behavior of the reservoir, i.e., the values in $\Wh$, $\Wx$ and $\mathbf{b}$ can be left untrained provided that the state dynamics satisfy a global asymptotic stability property called the Echo State Property (ESP) \cite{Jaeger2001, Yildiz2012}.
Essentially, the ESP states that when driven by a long input time-series, reservoir dynamics tend to synchronize \cite{verzelli2021learn} irrespectively of initial conditions, whose influence on the state progressively fade away.
Tailoring the reservoir initialization to the properties of the input signal and task still represents a challenging theoretical research question \cite{ceni2020echo, Manjunath2013}. However, in RC practice, it is common to follow a simple initialization process linked to a necessary condition for the ESP \cite{Lukosevicius2009, Jaeger2001, lukovsevivcius2012practical}. The recurrent reservoir weight matrix $\Wh$ is initialized randomly, e.g., from a uniform distribution over $[-1, 1]$, and then re-scaled to limit its spectral radius $\rho(\Wh)$ to a value smaller than $1$. Notice that this would require computing the eigenvalues of potentially large matrices, and thereby might result in a computationally demanding process. 
A more convenient procedure, in this case, can be obtained by using the circular law from random matrix theory, 
for instance generating random weights from a uniform distribution whose parameters depend on the desired value of the spectral radius,
as illustrated in \cite{gallicchio2019fastsr}.
The weight values in $\Wx$ and $\mathbf{b}$
are randomly drawn from a uniform distribution over $[-\omega_{x}, \omega_{x}]$ and $[-\omega_{b}, \omega_{b}]$, respectively. The values of $\alpha$, $\rho(\Wh)$, $\omega_{x}$ and $\omega_{b}$ are treated as hyper-parameters.

The trainable readout can be implemented in several forms.
In this paper, we focus on classification tasks and use a dense (linear) output layer that is fed by the last state computed by the reservoir when run on an input time-series.\\

\noindent
\textbf{ESP limitations --}
Regarding the ESP, it is worth noticing that while enabling a stable encoding robust to input perturbations, the required stability property gives a fading memory structure to the reservoir dynamics. As shown in \cite{Gallicchio2011NN}, the reservoir state space tends to organize in a suffix-based fashion, in agreement with the architectural Markovian bias of contractive RNNs \cite{Tino2007, Hammer2003}.
On the one hand, this means that ESNs are particularly well suited to solve tasks defined in compliance with this characterization, i.e., where the information in the suffix of the input time-series is dominant in determining the output \cite{Gallicchio2011NN}.
On the other hand, reservoir dynamics are constrained to progressively forget previous stimuli, making it difficult to propagate input information efficiently through many time-steps, as is might be required for example in time-series classification tasks.\\

\noindent
\textbf{Ring Reservoirs -- } While randomization plays an important role in RC, several studies in literature tried to find reservoir organizations that perform better than just random ones. A fruitful line of research deals with constraining the reservoir topology \cite{Strauss2012design} to get desirable algebraic properties of the recurrent weight matrix $\Wh$, especially based on an orthogonal structure \cite{white2004short, henaff2016recurrent}.
A simple yet effective way to achieve this goal consists in shaping the reservoir topology such that the reservoir neurons are arranged to form a cycle, or a \emph{ring}.
This specific architectural variant of ESNs has been shown to have noticeable advantages, e.g., in terms of the richness of the developed dynamics, longer memory, and performance on non-linear tasks \cite{verzelli2020input, tino2020dynamical, Rodan2010minimum}. 
In this paper, we thereby consider ESN based on ring reservoirs (R-ESN) as a competitive alternative to the standard vanilla setup.

\section{The Euler State Network Model}
\label{sec.model}

To introduce the proposed 
model, we start by considering the operation of a continuous-time recurrent neural system modeled by the following ODE:
\begin{equation}
\label{eq.ode}
\begin{array}{ll}
\h'(t) & = f(\mathbf{h}(t), \mathbf{x}(t)) \\
\\
& = \tanh(\Wh\;\h(t) + \Wx\;\x(t)+\mathbf{b})
\end{array}
\end{equation}
where
$\h(t)\in\R^N$ and $\x(t)\in\R^X$ respectively denote the state and the driving input, $\Wh \in \R^{N\times N}$ is the recurrent weight matrix, $\Wx \in \R^{N\times X}$ is the input weight matrix, and $\mathbf{b} \in \R^N$ is the bias.

We are interested in applying two types of constraints to the system in eq.~\ref{eq.ode}, namely \emph{stability} and \emph{non-dissipativity}. The former is required to develop a robust information processing system across time-steps, avoiding explosion of input perturbations that would result in poor generalization. The latter is important to avoid developing lossy dynamics, which would determine catastrophic forgetting of past inputs during the state evolution.

In what follows, in order to develop our model, we make use of results from stability theory of autonomous dynamical systems. As the system in eq.~\ref{eq.ode} is non-autonomous, the mathematical analysis is then approximated in the sense that stability and non-dissipativity can be interpreted as referred to small perturbations on the state of the system.
Notwithstanding this approximation, the mathematical analysis reported here is 
useful for deriving empirical initialization conditions in line with the RC literature \cite{Verstraeten2007}, as well as with the analysis presented in \cite{chang2019antisymmetricrnn} for fully trainable RNN models.

In the context of ODE stability, a fundamental role is played by the Jacobian of the system dynamics. In our case, the Jacobian of the system in eq.~\ref{eq.ode} is given by the following equation:
\begin{equation}
\label{eq.jacobian1}
\J_f(\h(t),\x(t)) = 
\frac{\partial f (\h(t),\x(t))}{\partial \h(t)} = 
\D(t) \, \Wh,
\end{equation}
in which $\D(t)$ is a diagonal matrix whose non-zero entries are given by $1-\tilde{\mathbf{h}}_1^2(t)$, $1-\tilde{\mathbf{h}}_2^2(t)$,$\ldots$,
$1-\tilde{\mathbf{h}}_N^2(t)$, with $\tilde{\mathbf{h}}_i(t)$ denoting the $i$-th component of the vector 
$\tanh\Big(\Wh \h(t) + \Wx \x(t) + \mathbf{b}\Big)$.

The interested reader can find a precise formulation of the stability of ODEs solutions in \cite{ascher1995numerical, glendinning1994stability}. Here, we limit ourselves to recall that
the solution to an ODE is stable if the real part of all eigenvalues of the corresponding Jacobian are $\leq 0$\footnote{In the context of non-autonomous ODEs, this condition comes down to a weakly contractive constraint on the system's dynamics.}. 
Using $\lambda_k(\cdot)$ to indicate 
the $k$-th eigenvalue of its matrix argument, the stability constraint in our case can be expressed as
\begin{equation}
\label{eq.stability}
\max_k {Re(\lambda_k(\J_f(\h(t),\x(t))))} \leq 0.
\end{equation}
Essentially, the real parts of the eigenvalues of the Jacobian indicate the degree of local ``expansion'' (or ``contraction'') of the state along specific dimensions of the space.
As discussed in \cite{haber2017stable, chang2019antisymmetricrnn}, 
having a $0$ upper bound on these quantities is important to avoid unbounded amplification of signal transmission.
On the other hand, if the real parts of these eigenvalues are $\ll 0$, then the resulting dynamics are lossy, in the sense that the state information is progressively shrunk exponentially rapidly over time\footnote{The solution of the linearized system evolves state perturbations proportionally to the exponential of the Jacobian multiplied by time \cite{glendinning1994stability}. The real parts of the eigenvalues of the Jacobian then determine the rate of exponential amplification (if positive) or shrinking (if negative) of state perturbations. In particular, when the real parts of such eigevalues is $\ll 0$, state perturbations quickly vanish and no longer have any influence on the dynamics.}. We thereby seek for a critical condition under which the eigenvalues of the Jacobian of our ODE 
are featured by $\approx 0$ real parts, i.e.:
\begin{equation}
\label{eq.condition}
Re(\lambda_k(\J_f(\h(t),\x(t)))) \approx 0
\quad k = 1, \ldots, N.
\end{equation}

A rather simple way to meet the condition in eq.~\ref{eq.condition} is to require that the recurrent weight matrix $\Wh$ in eq.~\ref{eq.ode} is antisymmetric \cite{haber2017stable}, 
i.e., it satisfies the condition:
\begin{equation}
\label{eq.antisymmetric}
\Wh = -\Wh^T.
\end{equation}
In fact, 
under the mild assumption that the diagonal matrix $\D(t)$ is invertible (i.e., in our case for neurons in non-saturating regime),
the eigenvalues of 
the Jacobian $\J_f(\h(t),\x(t)) = \D(t)\,\Wh$, are all purely imaginary\footnote{The eigenvalues of $\Wh$ are all purely imaginary due to its antisymmetric structure (see, e.g., \cite{trefethen1997numerical, strang1993introduction}). Using $\mathbf{S}$ to denote the square root of $\D(t)$, i.e., $\mathbf{S} = \D(t)^{1/2}$, it can be seen that $\D(t)\,\Wh$ is similar to $\mathbf{S}\,\Wh\,\mathbf{S}$ (a change of basis matrix is given by $\mathbf{P} = \D(t)^{-1/2}$), and thereby has its same eigenvalues. As $\mathbf{S}\,\Wh\,\mathbf{S}$ is itself antisymmetric, it follows that the eigenvalues of $\D(t)\,\Wh$ are all purely imaginary. The interested reader can find a different proof, under the same conditions, in \cite{chang2019antisymmetricrnn}.}.
Crucially, 
this property is \emph{intrinsic} to the antisymmetric structure imposed to the recurrent weight matrix $\Wh$, irrespectively of the particular choice of its values. In other words, the targeted critical dynamical behavior can be seen as 
an \emph{architectural bias} of the neural system described by eq.~\ref{eq.ode} under the imposed antisymmetric constraint in eq.~\ref{eq.antisymmetric}. Taking an RC perspective, we can then leave all the weight values in eq~\ref{eq.ode}, i.e., $\Wh$, $\Wx$ and $\mathbf{b}$, \emph{untrained}, provided that eq.~\ref{eq.antisymmetric} is satisfied.

Finally, to develop our discrete-time reservoir dynamics, we note that the ODE in eq.~\ref{eq.ode} can be solved numerically by using the forward Euler method \cite{suli2003introduction}, resulting in the following discretization:
\begin{equation}
\label{eq.F}
\begin{array}{ll}
\h(t) &= F(\h(t-1),\x(t))\\
\\
& = \h(t-1)\; + \\
          & \;\;\;
          \varepsilon\tanh
          \Big((\Wh-\gamma\I)\h(t-1)+\Wx\x(t)+\b\Big),
\end{array}
\end{equation}
where $\varepsilon$ is the step size of integration, and $\gamma$ is a diffusion coefficient used for stabilizing the discrete forward propagation \cite{haber2017stable}.
Note that both $\varepsilon$ and $\gamma$ are typically small positive scalars, and are treated as hyper-parameters.
We can now look at eq.~\ref{eq.F} as the state transition equation of a discrete-time \emph{reservoir} layer with $N$ recurrent neurons and $X$ input units, where $\Wh$, $\Wx$ and $\mathbf{b}$ are untrained, and $\Wh$ satisfies the constraint of being antisymmetric in eq.~\ref{eq.antisymmetric}. Since the untrained reservoir dynamics evolves as the forward Euler solution to an ODE, we call the resulting model the \emph{Euler State Network} (EuSN).
As in standard RC/ESN, 
the reservoir state is initialized at the origin, i.e., $\h(0) = \mathbf{0}$. Moreover,
the reservoir feeds a \emph{readout} layer, which is the only trained component of the EuSN architecture.

To \emph{initialize} the reservoir, we propose the following simple procedure, analogous to the common approaches in RC literature:
\begin{itemize}
    \item To setup the recurrent weights, we start by randomly initializing the values of a matrix $\W \in \R^{N\times N}$ from a uniform distribution over $[-\omega_{r}, \omega_{r}]$, where $\omega_{r}$ is a positive \emph{recurrent scaling} coefficient. We then set 
    $\Wh = \W - \W^T$, which satisfies eq.~\ref{eq.antisymmetric} as it is antisymmetric by construction. 
    \item The values of the input weight matrix $\Wx$ are randomly initialized from a uniform distribution over $[-\omega_{x}, \omega_{x}]$, where $\omega_{x}$ is a positive \emph{input scaling coefficient} as in ESNs.
    \item The values of the bias vector $\mathbf{b}$ are randomly initialized from a uniform distribution over $[-\omega_{b}, \omega_{b}]$, where $\omega_{b}$ is a positive \emph{bias scaling coefficient} as in ESNs.
\end{itemize}

Notice that the stability properties of the reservoir are due to the antisymmetric structure of $\Wh$, without requiring to scale its spectral radius as in ESNs. The recurrent, input and bias scaling coefficients $\omega_{r}$, $\omega_{x}$, and $\omega_{b}$, are introduced simply to balance the contributions of the different terms in the state transition eq.~\ref{eq.F}, and are treated as hyper-parameters. 

It is worth mentioning that, in the broader landscape of trainable RNN research, the introduced EuSN model brings similarities to the Antisymmetric RNN (A-RNN) \cite{chang2019antisymmetricrnn}, which is featured by state dynamics evolving as
in eq. \ref{eq.F}, under a similar antisymmetric constraint on the recurrent weight matrix.
Crucially, while in EuSN all the internal weights
are left untrained after initialization, in A-RNN they are all trainable. In this view, the study of EuSN can be seen under the perspective of randomized neural networks \cite{gallicchio2020deep}, emphasizing the intrinsic capabilities of stable recurrent architectures based on forward Euler discretization and antisymmetric recurrent weight matrices even in the absence (or prior to) training of internal connections.

The following Section~\ref{sec.analysis} delves into the mathematical analysis of the EuSN reservoir system, studying its asymptotic stability properties and its intrinsically critical dynamical regime.

\section{Mathematical Analysis}
\label{sec.analysis}
We study the reservoir of an EuSN
as a discrete-time dynamical system evolving according to the function $F$ defined in eq.~\ref{eq.F},
where we remind that $\Wh$ is antisymmetric, and both $\varepsilon$ and $\gamma$ are small positive reals.
We use here the same mathematical approximations used in Section~\ref{sec.model} to define the model.
The local behavior of the reservoir system around a specific state $\h_0$ can be analyzed by linearization, i.e., by approximating its dynamics as follows:
\begin{equation}
\label{eq.linearized}
\h(t) = \J_{F}(\h_0,\x(t))\;(\h(t-1) - \h_0) + F(\h_0,\x(t)).
\end{equation}
Here $\J_{F}(\h_0,\x(t))$ is used to denote the Jacobian of the function $F$ in eq.~\ref{eq.F} at time-step $t$ and evaluated at $\h_0$, i.e.:
\begin{equation}
\label{eq.Jacobian}
\J_{F}(\h_0,\x(t)) = \frac{\partial F (\h(t-1),\x(t))}{\partial \h(t-1)}\Bigr|_{\h = \h_0}.
\end{equation}

In classical ESN literature it is common to study the eigenspectrum of the Jacobian of the reservoir, and especially its spectral radius, in order to characterize the asymptotic stability behavior of the system. In particular, when introducing necessary conditions for the ESP \cite{Jaeger2001}, 
it is common to study the autonomous system (i.e., in the absence of driving input $\x(t) = \mathbf{0}$ for every $t$, and bias $\mathbf{b} = \mathbf{0}$), linearized around  the origin (i.e., $\h_0 = \mathbf{0}$). For ESNs, this procedure then results in an algebraic constraint that is imposed on the recurrent weight matrix of the reservoir in order to control its effective spectral radius.
Interestingly, by taking the same assumptions in the case of EuSN, we find that all the eigenvalues of the Jacobian are naturally confined 
to values close to 1. 
More precisely, the following Proposition~\ref{th.proposition1} characterizes the values that can be taken by the eigenvalues of the Jacobian of the autonomous EuSN reservoir system linearized around the origin. In the following, we use the notation $\lambda_k(\cdot)$ to indicate the $k$-th eigenvalue of its argument.

\begin{proposition}
\label{th.proposition1}
Let us consider an EuSN whose reservoir dynamics are ruled by the state transition function in eq. \ref{eq.F}, with step size $\varepsilon$, diffusion coefficient $\gamma$, and reservoir weight matrix $\Wh$.
Assume autonomous (i.e., with null driving input and bias) and linearized dynamics around the origin. Then, the eigenvalues of the resulting Jacobian satisfy the following condition:
\begin{equation}
\label{eq.eigs}
\begin{array}{l}
\text{for } k = 1, \ldots, N: 
\vspace{1mm}
\\
Re\big(\lambda_k(\J_F(\mathbf{0},\mathbf{0}))\big) = 1-\varepsilon\, \gamma,
\vspace{1mm}
\\
Im\big(\lambda_k(\J_F(\mathbf{0},\mathbf{0}))\big)
\in [-\varepsilon\, \rho(\Wh),\, \varepsilon\, \rho(\Wh)],
\end{array}
\end{equation}
\begin{proof}
Computing an expression for the Jacobian in eq.~\ref{eq.Jacobian} around the origin and in the absence of external input, we find that:
\begin{equation}
\label{eq.J00}
\J_F(\mathbf{0},\mathbf{0}) = \I + \varepsilon (\Wh - \gamma \I) = (1-\varepsilon\,\gamma) \I +\varepsilon\, \Wh.
\end{equation}
We can notice that each eigenvalue of the sum of matrices in the right-hand side of eq.~\ref{eq.J00} is given by an eigenvalue of $\I (1-\varepsilon\gamma)$ added to an eigenvalue of $\varepsilon \Wh$
(in facts, the two matrices commute). Thereby, we have:
\begin{equation}
\label{eq.lambdak}
\lambda_k(\J_F(\mathbf{0},\mathbf{0})) = 1-\varepsilon\,\gamma + \varepsilon\, \lambda_k(\Wh) \quad k = 1, \ldots, N.
\end{equation}
As matrix $\Wh$ is antisymmetric, its eigenvalues are purely imaginary. In particular, except for a $0$ eigenvalue in correspondence to the odd-dimensional case, they come in the form 
$\lambda_k(\Wh) = i\, \beta_k$, where $i$ is used to denote the imaginary unit, and $\beta_k = Im(\lambda_k(\Wh))$ can take the possible values $\pm b_1$, $\ldots$, $\pm b_{\left \lfloor{N/2}\right \rfloor}$.
We can then rewrite eq.~\ref{eq.lambdak} as follows:
\begin{equation}
\label{eq.lambdak2}
\lambda_k(\J_F(\mathbf{0},\mathbf{0})) = 
1-\varepsilon\, \gamma + i \; \varepsilon \, \beta_k
\quad k = 1, \ldots, N,
\end{equation}
from which it is straightforward to conclude that $Re\big(\lambda_k(\J_F(\mathbf{0},\mathbf{0}))\big) = 1-\varepsilon \, \gamma$.
Moreover, observing that $\beta_k \in [-\rho(\Wh),\, \rho(\Wh)]$, we can also derive that
$Im\big(\lambda_k(\J_F(\mathbf{0},\mathbf{0}))\big) \in [-\varepsilon\,\rho(\Wh),\, \varepsilon\,\rho(\Wh)]$.

\end{proof}
\end{proposition}

Proposition~\ref{th.proposition1} enables us to locate the eigenvalues of $\J_F(\mathbf{0},\mathbf{0})$ as a function of $\varepsilon$, $\gamma$ and $\rho(\Wh)$.
Recalling that, in practice, both $\varepsilon$ and $\gamma$ are small positive scalars, we can observe that 
$\lambda_k(\J_F(\mathbf{0},\mathbf{0}))\approx 1$ for $k = 1, \ldots, N$.
This
has interesting consequences in terms of the asymptotic behavior of the reservoir. Indeed, as all the eigenvalues of the Jacobian $\J_F(\mathbf{0},\mathbf{0})$ are $\approx 1$, it follows that the autonomous operation of the reservoir 
system around the origin tends to preserve all the components of the state, while performing infinitesimal rotations due to the antisymmetric structure of the recurrent matrix. Any perturbation to the zero state 
(i.e., of the origin) is preserved over time without vanishing, which
essentially means that EuSNs do not tend to show the ESP under the proposed initialization conditions.

\begin{figure*}
     \centering
     \begin{subfigure}[b]{0.4\textwidth}
         \centering
         \includegraphics[width=\textwidth]{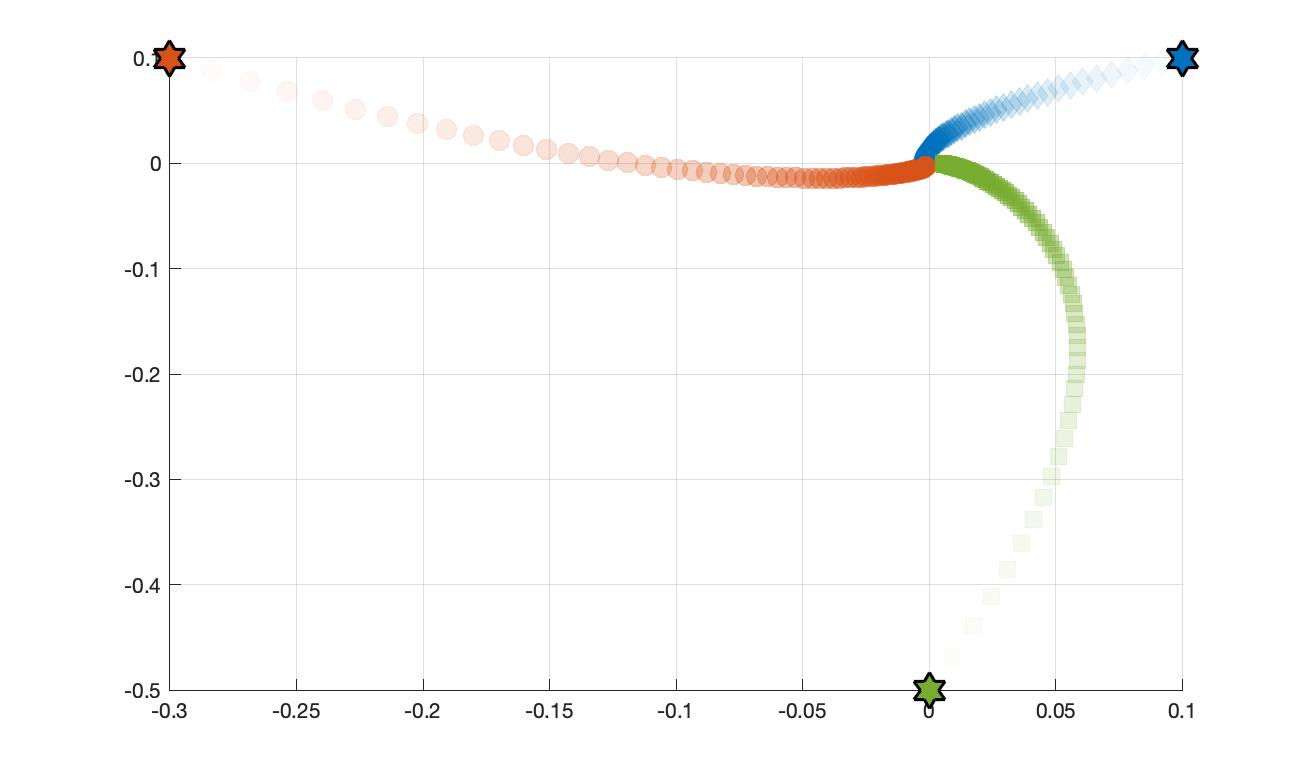}
         \caption{ESN with ESP.}
         \label{fig.states_esn_esp}
     \end{subfigure}
     \begin{subfigure}[b]{0.4\textwidth}
         \centering
         \includegraphics[width=\textwidth]{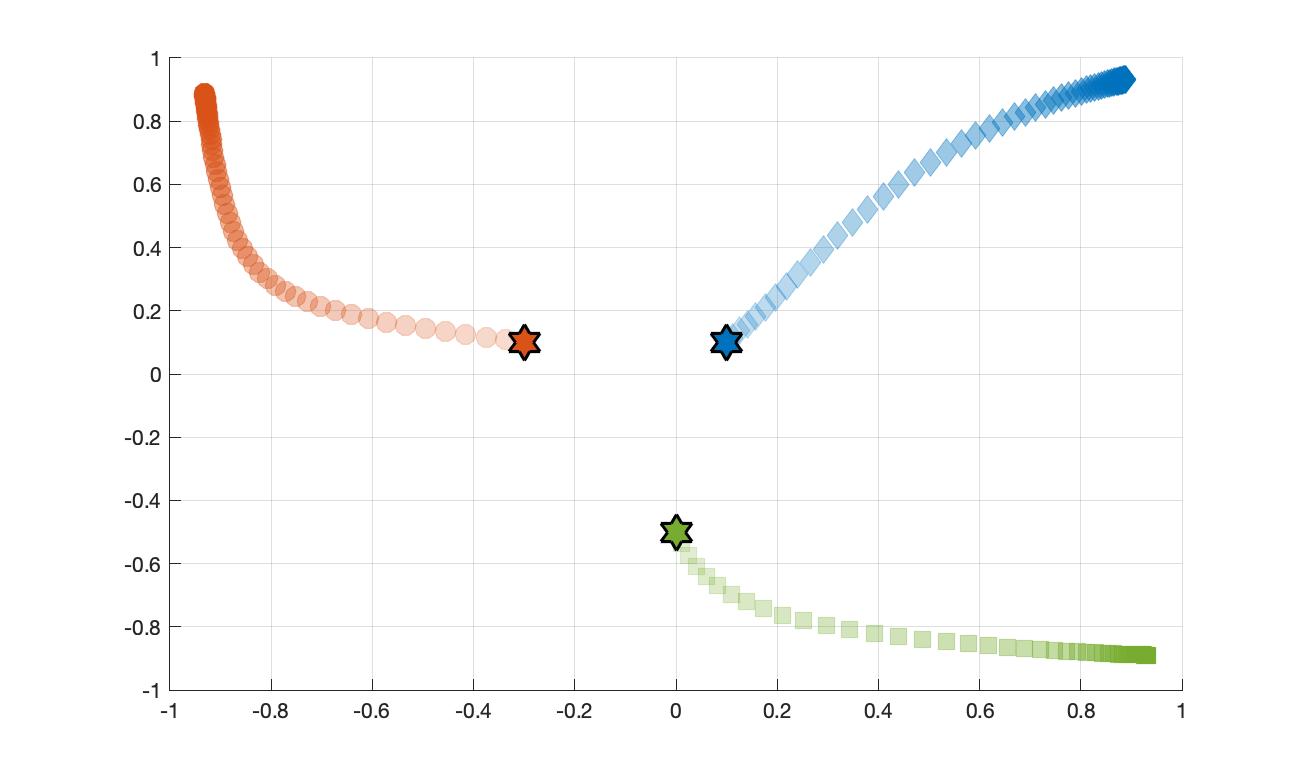}
         \caption{ESN without ESP.}
         \label{fig.states_esn_noesp}
     \end{subfigure}
     \begin{subfigure}[b]{0.4\textwidth}
         \centering
         \includegraphics[width=\textwidth]{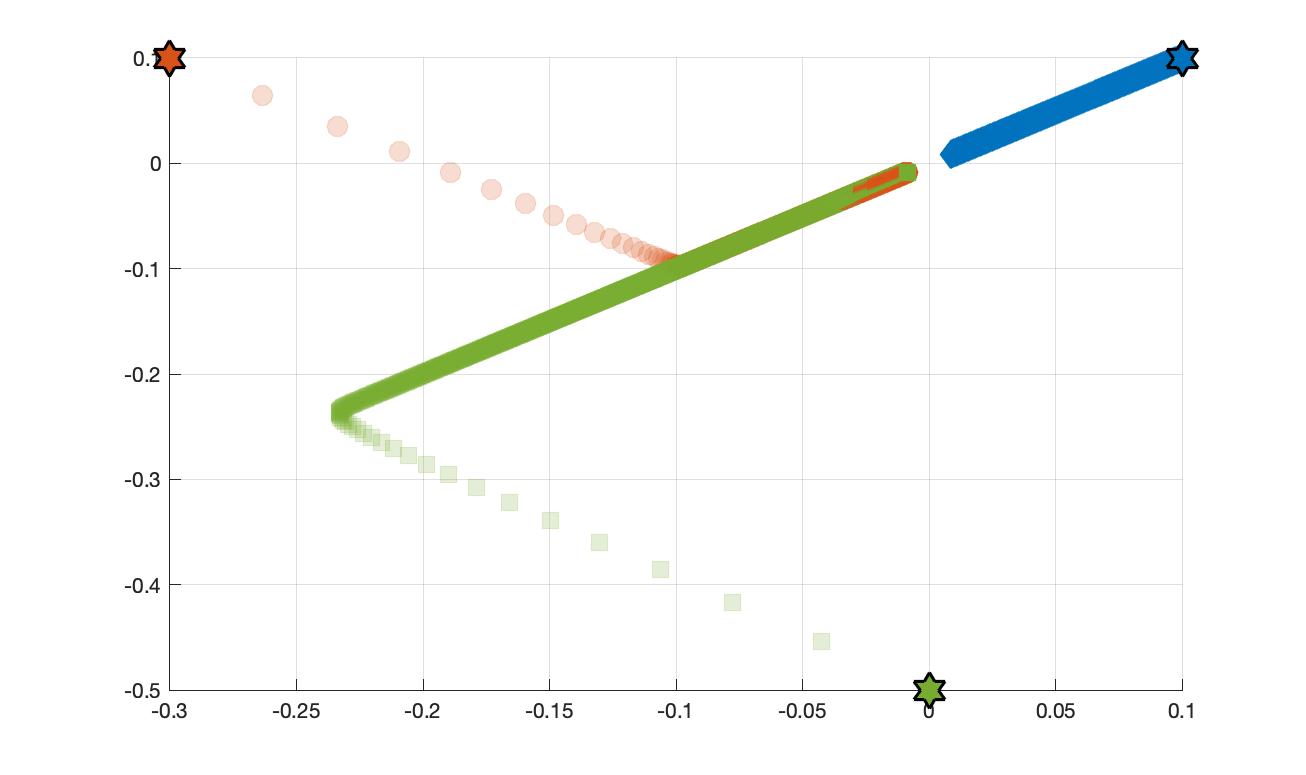}
         \caption{R-ESN.}
         \label{fig.states_resn}
     \end{subfigure}
     \begin{subfigure}[b]{0.4\textwidth}
         \centering
         \includegraphics[width=\textwidth]{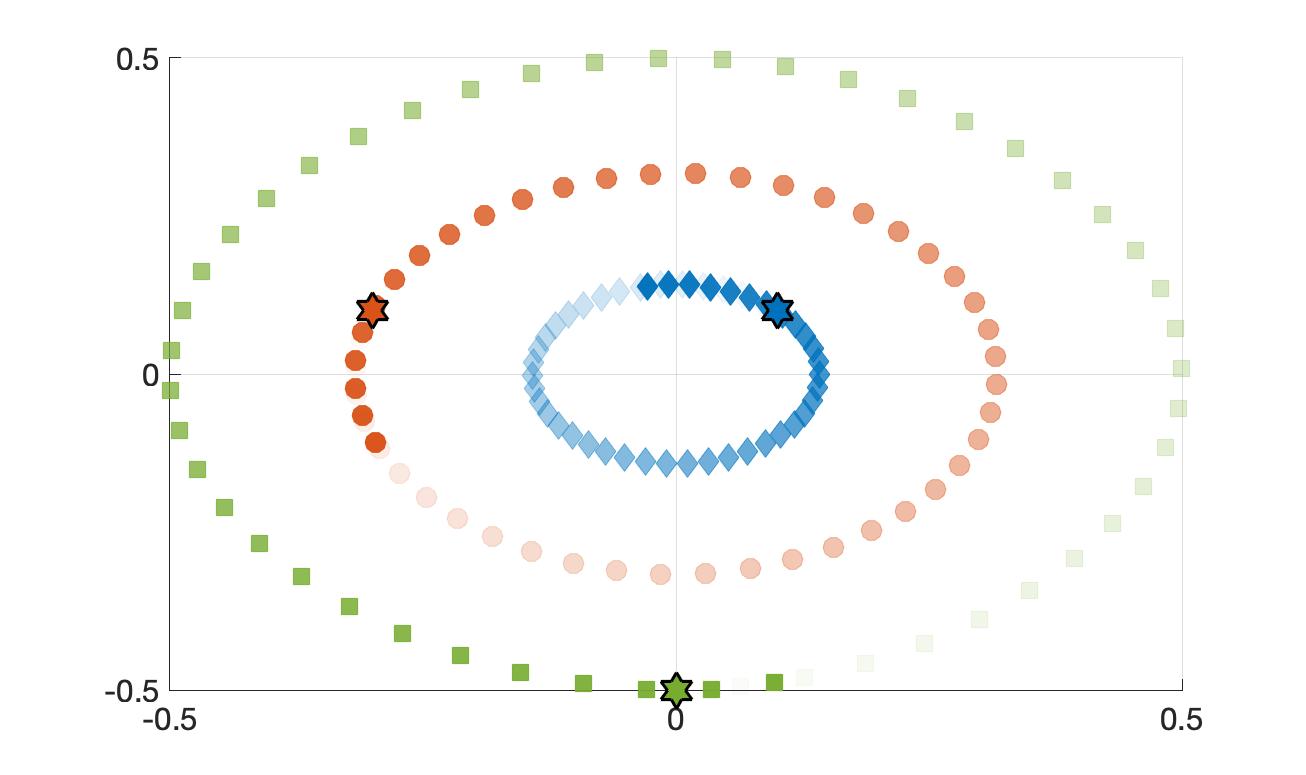}
         \caption{EuSN.}
         \label{fig.states_eusn}
     \end{subfigure}
     \caption{Autonomous reservoir dynamics 
     around the origin visualized for 
     instances of: (a) ESN with ESP, (b) ESN without ESP, 
     (c) R-ESN, and (d) EuSN. The same three initial conditions are used in all the cases, illustrated as full stars of different colors: $\mathbf{u}_0 = [0.1, 0.1]^T$ in blue, $\mathbf{v}_0 = [-0.3, 0.1]^T$ in red, and $\mathbf{z}_0 = [0, -0.5]^T$ in green.
     Trajectories in the 2-dimensional reservoir space are shown by (sampled) points of different shapes, where more transparent colors indicate earlier time-steps. 
     The recurrent weight matrix is: $\Wh = \big[[0.7,0.1]^T , [-0.1,0.7]^T \big]$ in (a); $\Wh = \big[[1.7,0.1]^T , [-0.1,1.7]^T \big]$ in (b); $\Wh = \big[[0,1]^T , [1,0]^T \big]$ in (c); $\Wh = \big[[0,1.5]^T , [-1.5,0]^T \big]$ in (d).
     In (a), (b), and (c) the leaking rate is $\alpha = 0.001$. In (d) the step size and the diffusion coefficient are $\varepsilon, \gamma = 0.001$.}
     \label{fig.states}
\end{figure*}

The qualitative difference between the behaviors of an EuSN and 
those of an ESN is graphically illustrated in 
Figure~\ref{fig.states}. In the figure, taking inspiration from the graphical analysis of the inner dynamics of deep architectures used in \cite{haber2017stable} and \cite{chang2019antisymmetricrnn}, we visualize
examples of state trajectories developed by 2-dimensional autonomous 
reservoirs in different conditions: ESN with and without ESP, R-ESN with orthogonal weight matrix, and EuSN.
In the case of ESNs, when the ESP is valid (Figure~\ref{fig.states_esn_esp}), the origin is an attractor and all the trajectories 
asymptotically
converge to it.
When instead the ESN does not satisfy the ESP (Figure~\ref{fig.states_esn_noesp}), the origin is a repeller.
In the case of R-ESN (Figure~\ref{fig.states_resn}), even if the recurrent weight matrix is orthogonal, trajectories still converge to the origin (more slowly than in Figure~\ref{fig.states_esn_esp}) due to contractive properties of the $\tanh$ non-linearity. Interestingly, in the case of EuSN (Figure~\ref{fig.states_eusn}), the states rotate around the origin without being attracted nor repelled, and the distance between the orbits are preserved, reflecting the differences among the initial conditions. In other words\footnote{The different initial conditions can be interpreted as the same reservoir state perturbed by different external input values.}, differently from ESN variants, the effect of the external input on the reservoir dynamics in EuSN does not die out nor explode with time.

A closer inspection of eq.~\ref{eq.lambdak2} also reveals that the small deviations from $1$ in the eigenvalues of the Jacobian in $\J_F(\mathbf{0},\mathbf{0})$ are possibly due to the combined effects of the actual values of $\varepsilon$, $\gamma$, and of the 
imaginary part of the
eigenvalues of $\Wh$.
In this case, it is also possible to derive a simple and direct expression for the effective spectral radius of the autonomous linearized EuSN system, as given by the following Corollary~\ref{th.corollary}.

\begin{corollary}
\label{th.corollary}
Let us consider an EuSN whose reservoir dynamics are ruled by the state transition function in eq. \ref{eq.F}, with step size $\varepsilon$, diffusion coefficient $\gamma$, and reservoir weight matrix $\Wh$.
Assume autonomous (i.e., with null driving input and bias) and linearized dynamics around the origin. The spectral radius of the resulting Jacobian is then given by the following expression:
\begin{equation}
\label{eq.spectralradius_eusn}
\rho(\J_F(\mathbf{0},\mathbf{0})) = \sqrt{1+\varepsilon^2\gamma^2
                                    -2\varepsilon\gamma+
                                    \varepsilon^2\rho(\Wh)^2}.
\end{equation}
\end{corollary}
\begin{proof}
From Proposition~\ref{th.proposition1}, we know that 
the eigenvalues of $\J_F(\mathbf{0},\mathbf{0})$ with the largest modulus are 
$1-\varepsilon\,\gamma \pm i\, \varepsilon \, \rho(\Wh)$.
From this observation, the spectral radius of the Jacobian is easily computed, as follows:
\begin{equation}
\label{eq.rhoJ}
\begin{array}{ll}
\rho(\J_F(\mathbf{0},\mathbf{0})) 
& = 
\sqrt{(1 + \varepsilon \,\gamma)^2 + (\varepsilon\, \rho(\Wh))^2}
\vspace{1mm}
\\
& =
\sqrt{1 + \varepsilon^2\gamma^2 - 2\varepsilon\gamma + \varepsilon^2 \rho(\Wh)^2}.
\end{array}
\end{equation}
\end{proof}

Interestingly, applying the result of Corollary~\ref{th.corollary} in correspondence of small values of both $\varepsilon$ and $\gamma$, we can easily observe that $\rho(\J_F(\mathbf{0}, \mathbf{0})) \approx 1$ by design. In particular, 
Corollary~\ref{th.corollary} clearly indicates the marginal role of the spectral radius of the recurrent weight matrix $\Wh$ in determining the 
effective spectral radius of the EuSN, hence the asymptotic behavior of the system. The influence of $\rho(\Wh)$ on $\rho(\J_F(\mathbf{0}, \mathbf{0}))$ is indeed progressively more negligible for smaller values of the step size parameter $\varepsilon$. This aspect represents another crucial difference with respect to ESNs, and gives support to the initialization strategy proposed in Section~\ref{sec.model}, which does not require scaling the spectral radius of $\Wh$, and rather suggests a simpler scaling by an $\omega_{r}$ hyper-parameter after the random initialization. 

\begin{figure*}
     \centering
     \includegraphics[width=0.6\textwidth]{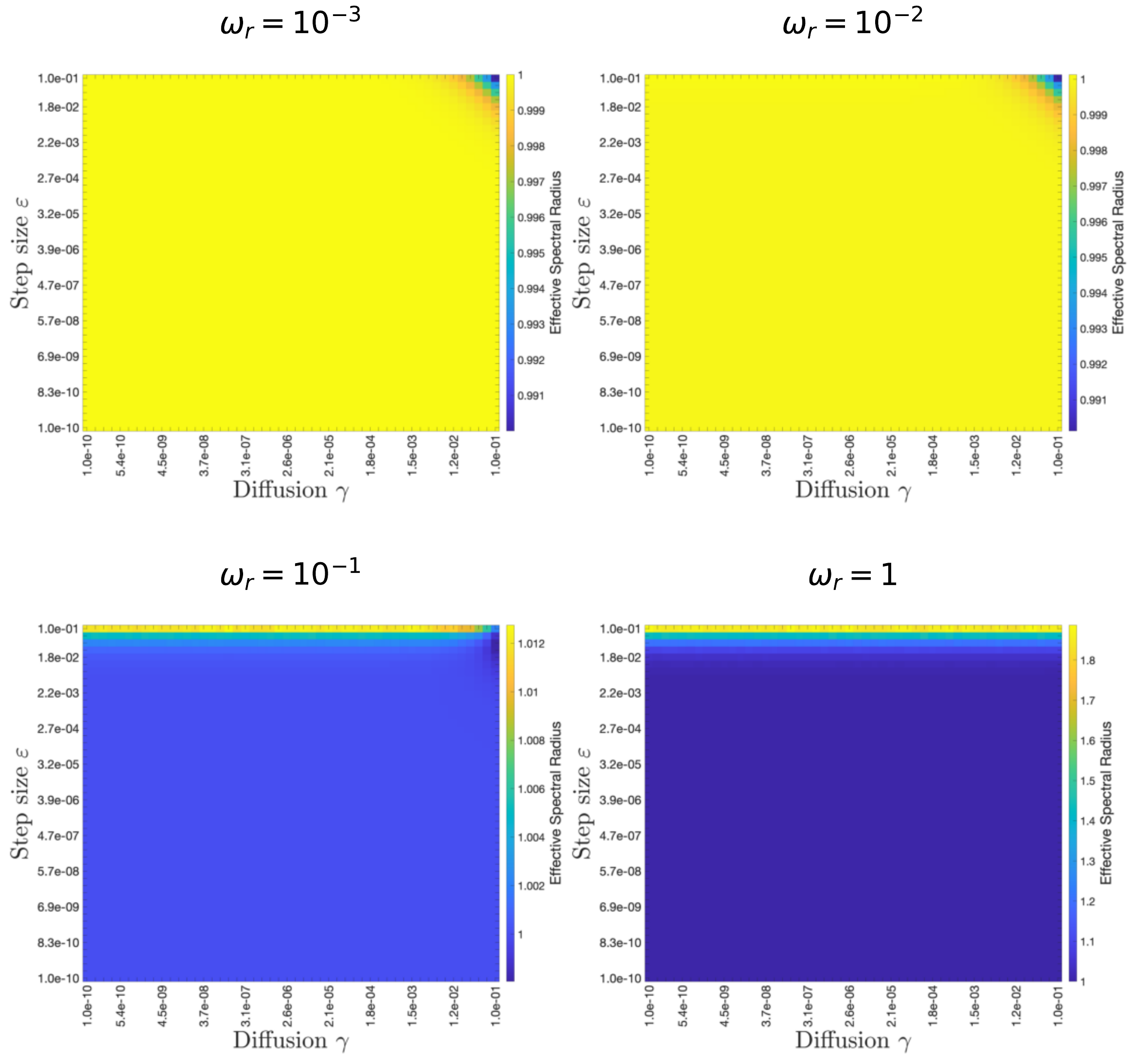}
     
     \caption{Effective spectral radius of EuSN with $N = 100$ reservoir neurons, for varying values of the step size ($\varepsilon$, vertical axis in each plot),
     of the diffusion ($\gamma$, horizontal axis in each plot), and of scaling of the recurrent weight matrix $\Wh$ ($\omega_{r}$, different plots). For each configuration, the results are averaged over 3 repetitions. }
     \label{fig.rho}
\end{figure*}

Figure~\ref{fig.rho} provides some examples of values of the effective spectral radius of EuSN, computed considering reservoirs with $N = 100$ neurons, varying $\varepsilon$ and $\gamma$ between $10^{-10}$ and $10^{-1}$, and for different values of $\omega_{r}$ varying from $0.001$ to $1$ (with resulting values of $\rho(\Wh)$ ranging from $\approx 0.015$
to $\approx 15$). For every configuration, the reported values are computed by averaging over 3 repetitions,
following the 
reservoir initialization process described in Section~\ref{sec.model}. Results confirm that the 
effective
spectral radius is restricted to values $\approx 1$ in all the configurations, with a few exceptions, represented by the largest values of $\varepsilon$ and $\omega_{r}$.

We now extend our analysis to conditions that include non-zero input signals. To this end, we find it useful to consider the spectrum of (pseudo) local Lyapunov exponents (LLEs) \cite{bailey1996, abarbanel1992local} of the reservoir. 
An $N$-dimensional system has $N$ LLEs, denoted as $LLE_1, \ldots, LLE_N$. 
These quantities provide a measure of the sensitivity of the system to small perturbations of the state trajectories, characterizing the dynamical behavior locally to the state evolved over time. In particular, the maximum local Lyapunov exponent (MLLE) 
is a useful indicator of the regime of the reservoir dynamics
\cite{Verstraeten2007,Verstraeten2009,Bianchi2016}
:
values $>0$ indicate unstable dynamics, values $< 0$ denote stable dynamics, while the condition of $MLLE = 0$ corresponds to a \emph{critical} condition of transition between stability and instability, often referred to as the \emph{edge of stability} (or the edge of chaos). In the RC literature, it has been observed that reservoir systems operating at the edge of stability develop richer dynamics and often result in higher performance in applications \cite{gallicchio2018local, legenstein2007makes, Boedecker2012}.
%
For the case of EuSN, we can find interesting bounds for all the LLEs (hence also for the MLLE), as stated by the following Proposition~\ref{th.proposition2}.

\begin{proposition}
\label{th.proposition2}
Let us consider an EuSN whose reservoir dynamics are ruled by the state transition function in eq. \ref{eq.F}, with step size $\varepsilon$, diffusion coefficient $\gamma$, and reservoir weight matrix $\Wh$.
The LLEs of the reservoir satisfy the following condition:
\begin{equation}
\label{eq.prop_lle}
\begin{array}{l}
\text{for } k = 1, \ldots, N: \\
LLE_k \in [\ln\big(1-\varepsilon(\rho(\Wh)+\gamma)\big), \ln\big(1+\varepsilon(\rho(\Wh)+\gamma)\big)].
\end{array}
\end{equation}
\begin{proof}
Suppose to drive the reservoir dynamics by an external 
time-series of length $T$, i.e., $\x(1), \ldots \x(T)$.
To evaluate the LLEs, we consider the dynamics along the resulting state trajectory and apply the following estimator \cite{Bianchi2016}, for each $k = 1, \ldots, N$:
\begin{equation}
\label{eq.LLE}
LLE_k =
\frac{1}{T}
\sum_{t = 1}^T
\ln |\lambda_k(\J_F(\h(t-1),\x(t)))|.
\end{equation}
Computing the full Jacobian in eq.~\ref{eq.Jacobian}, we get:
\begin{equation}
\label{eq.Jacobian_full}
\J_F(\h(t-1),\x(t)) =
\I + \varepsilon \mathbf{D}(t) \Wh - 
\varepsilon \gamma \mathbf{D}(t),
\end{equation}
where $\mathbf{D}(t)$ is a diagonal matrix 
with diagonal values given by $1-\tilde{\mathbf{h}}_1^2(t)$, $1-\tilde{\mathbf{h}}_2^2(t)$,$\ldots$,
$1-\tilde{\mathbf{h}}_N^2(t)$, with $\tilde{\mathbf{h}}_i(t)$ denoting the $i$-th component of the vector 
$\tanh\Big((\Wh-\gamma\I)\h(t-1) + \Wx \x(t) + \mathbf{b}\Big)$.

By virtue of the Bauer-Fike's theorem \cite{bauer1960norms}, we can observe that the eigenvalues of $\J_F(\h(t-1),\x(t))$ satisfy the following inequality:
\begin{equation}
\label{eq.bauer}
\begin{array}{ll}
\text{for }k = 1,\ldots, N:& \\
|\lambda_k(\J_F(\h(t-1),\x(t))) - 1| & \leq 
\|\varepsilon \D(t) \Wh - \varepsilon \gamma \D(t)\|_2 \\
& = \varepsilon \| \D(t) \Wh - \gamma \D(t)\|_2 \\
& \leq \varepsilon \|\Wh\|_2 + \varepsilon \gamma \\
& = \varepsilon (\rho(\Wh) + \gamma),
\end{array}
\end{equation}
where we have used that $\mathbf{D}(t)$ is a diagonal matrix with non-zero values in $[0,1]$, and $\|\Wh\|_2 = \rho(\Wh)$ for antisymmetric matrices. Eq.~\ref{eq.bauer} indicates that every eigenvalue of the Jacobian is bounded within a ball of radius $\varepsilon (\rho(\Wh) + \gamma)$ around $1$, and its modulus is thereby bounded as follows:
\begin{equation}
\label{eq.bound_abs}
\begin{array}{l}
\text{for } k = 1,\ldots,N:\\
1-\varepsilon (\rho(\Wh) + \gamma) \leq |\lambda_k(\J_F(\h(t-1),\x(t)))| 
\leq 1+ \varepsilon (\rho(\Wh) + \gamma).
\end{array}
\end{equation}
Finally, using eq.~\ref{eq.LLE} and the result from eq.~\ref{eq.bound_abs},
we can conclude that the LLEs of the reservoir satisfy the following condition:
\begin{equation}
\label{eq.bound_lle}
\begin{array}{l}
\text{for } k = 1,\ldots,N:\\
\ln\big(1-\varepsilon (\rho(\Wh) + \gamma)\big) 
\leq LLE_k \leq 
\ln\big(1+ \varepsilon (\rho(\Wh) + \gamma)\big),
\end{array}
\end{equation}
which gives the thesis.

\end{proof}
\end{proposition}


The result of Proposition~\ref{th.proposition2} allows us to bound the values of the LLEs of the reservoir based on $\varepsilon$, $\gamma$ and $\rho(\Wh)$.
Interestingly, as in practice both $\varepsilon$ and $\gamma$ are small positive scalars, the role of $\rho(\Wh)$ is negligible, and the result is that LLEs are all confined to a small neighborhood of $0$ due to the peculiar architectural design of the reservoir. In particular, $MLLE \approx 0$, meaning that the EuSN model is architecturally biased towards the critical dynamical regime near the edge of stability.

\begin{figure*}
     \centering
     \includegraphics[width=0.6\textwidth]{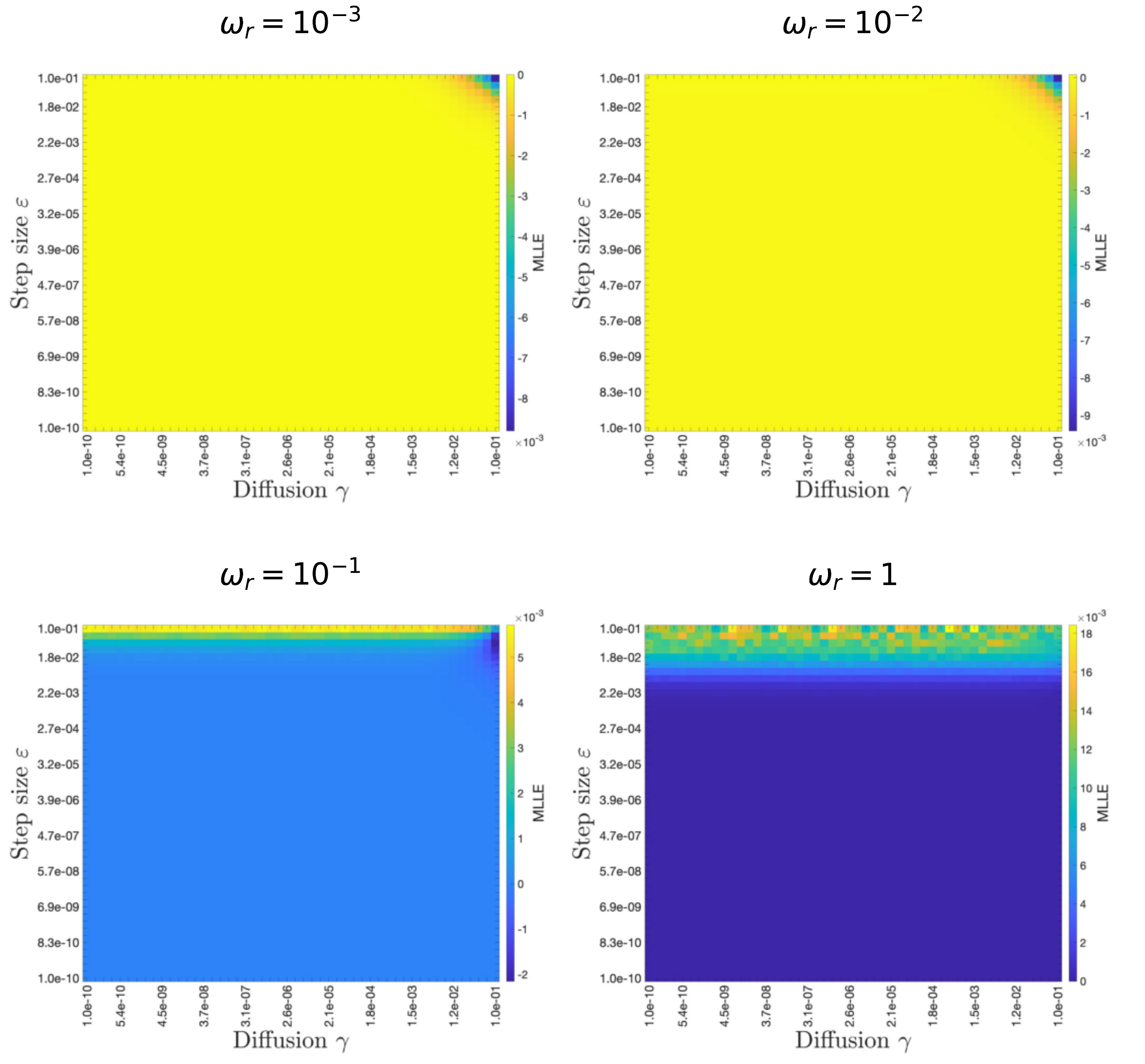}
     
     \caption{MLLE of EuSN with $N = 100$ reservoir neurons, for varying values of the step size ($\varepsilon$, vertical axis in each plot),
     of the diffusion ($\gamma$, horizontal axis in each plot), and of scaling of the recurrent weight matrix $\Wh$ ($\omega_{r}$, different plots). For each configuration, the results are averaged over 3 repetitions. See details in the text.}
     \label{fig.MLLE}
\end{figure*}

To concretely illustrate the result of Proposition~\ref{th.proposition2}, we numerically compute the LLEs of reservoir dynamics in EuSN with $N = 100$, using the estimator in eq.~\ref{eq.LLE}, and considering as driving input a one-dimensional time-series of length $500$ whose elements are randomly drawn from a uniform distribution over $[-0.5, 0.5]$, 
with $\omega_{x} = \omega_b = 1$.
%
Varying the $\varepsilon$, $\gamma$ and $\omega_{r}$ hyper-parameters as in Figure~\ref{fig.rho}, in Figure~\ref{fig.MLLE} we show the resulting MLLE (averaged over 3 repetitions for every configuration). As results clearly show, the obtained MLLE is $\approx 0$ in every configuration, confirming the analytical outcomes 
regarding the intrinsic critical regime of EuSN dynamics.

\section{Experiments}
\label{sec.experiments}
The mathematical analysis provided in Section~\ref{sec.analysis} pointed out that EuSN dynamics are naturally biased towards the edge of stability.
In this section, we show the relevant impact that such a characterization can have in applications.
First, in Section~\ref{sec.memory}, we analyze the impact on tasks involving long-term memorization in comparison to RC models. 
Then, in Section~\ref{sec.benchmarks} we analyze the resulting performance (in terms of accuracy, required times, and energy consumption) on several time-series classification benchmarks, in comparison to both RC models and trainable RNN architectures.

The experiments described in this section were run in single-GPU mode on a system equipped with 2x20 Intel(R) Xeon(R) CPU E5-2698 v4 @ 2.20GHz and 4 Tesla P100-PCIE-16GB GPU.
The code used for our experiments is written in Keras and Scikit-learn, and is made publicly available online\footnote{ 
\url{https://github.com/gallicch/EulerStateNetworks}}.

\subsection{Long-term Memorization}
\label{sec.memory}





We introduce a family of time-series classification tasks to assess the long-term memorization (LTM) capabilities of recurrent layers.
The general aim is to exercise the ability of neural networks to correctly classify an increasingly long input time-series based on the presence of specific patterns injected into the sequence at arbitrary points in time. The recurrent layer must be able to effectively latch input information into the state representations over long time spans, as the information relevant to the target becomes increasingly distant from the input suffix, posing a relevant challenge for RC-based systems. \cite{Gallicchio2011NN}.

We designed LTM experiments with 4 tasks.
The first one is based on a \emph{Synthetic} dataset. Specifically, we created two sequential patterns, $p_1$ and $p_2$, of $10$ time-steps each, and both containing 1-dimensional signals randomly sampled from a normal distribution. Each time-series for the task was obtained by first generating a prefix with 1-dimensional normal noise of random length, uniformly drawn between $0$ a $20$. Then, one of two patterns (either $p_1$ or $p_2$) was added to the prefix, and then a suffix with normal noise was added until a total length of $T$ was reached, varying $T$ from $30$ to $400$.
Half of the input time-series contain the first pattern $p_1$, and they are assigned the target class $+1$. The remaining time-series contain the second pattern $p_2$, and they are assigned the target class $0$.
For every choice of the total length of the padded sequences $T$, we generated a dataset of 1000 time-series, equally divided (with stratification) into training set and test set.
Data intended for training have been divided into a training and a validation set by a further (stratified) 50\%-50\% split.
The other tasks were obtained similarly, as padded versions of
\emph{LSST}~\cite{allam2018photometric}, \emph{PenDigits}~\cite{alimoglu1997combining}, and \emph{RacketSports}, a selection of classification benchmarks from the UEA \& UCR time-series classification repository \cite{timeseriesdatasets}, featured by short input time-series. In these cases, the original time-series from the respective datasets were used as injected patterns into longer sequences constructed as for the Synthetic task, with the difference that the random prefix and suffix had the same dimensionality as the original input signals.
Each time series constructed in this way was associated with the same target class of the corresponding sample in the original dataset. For each task, we used the original training-test splitting, with a further $67\% - 33\%$ training-validation stratified splitting. 
Finally, notice that for all the tasks, the difficulty in the classification is increased by design for larger values of the total length of padded sequences (i.e., $T$), since the input pattern relevant to the target is positioned in a progressively more distant past.

Table~\ref{tab.datasets_memorization} summarizes the main characteristics of the datasets used in the LTM experiments. \\

\begin{table*}[t]
\centering
\footnotesize
\begin{tabular}{lrrrrrr}
\hline
\textbf{Name}& 
\textbf{\#Seq Tr}&
\textbf{\#Seq Ts}&
\textbf{Pattern Length}&
\textbf{Padded Length}&
\textbf{Feat.}&
\textbf{Classes}\\
\hline
Synthetic & 500 & 500 & 10 & 30-400 & 1 & 2 \\
LSST & 2459 & 2466 & 36 & 60-400 & 6 & 14 \\
PenDigits & 7494 & 3498 & 8 & 30-400 & 2 & 10 \\
RacketSports & 151 & 152 & 30 & 50-400 & 6 & 4 
\end{tabular}
\caption{Information on the time-series classification tasks used for the experiments on LTM, namely:
size of the training (\#Seq Tr) and of the test set (\#Seq Ts); length of the pattern relevant to the target classification, i.e., length of the original time-series for LSST, RacketSports, and PenDigits (Pattern Length); min-max total length of the padded time-series $T$ (Padded Length); 
number of input features, i.e., input dimensionality $X$ (Feat.); number of class labels (Classes).}
\label{tab.datasets_memorization}
\end{table*}

\noindent
\textbf{Experimental Settings --}
We considered EuSNs with $N$
reservoir neurons between $10$ and $200$,
exploring values of the other network hyper-parameters in the following ranges:
$\omega_{r}$, $\omega_{x}$ and $\omega_{b}$ in $\{10^{j} :  j = -3, \ldots, 1\}$;
$\varepsilon$ and $\gamma$ 
in $\{10^{j} : j = -5, \ldots, 0\}$.
For comparison, we ran experiments with ESNs and R-ESNs configured with the same number of reservoir neurons. In this case, we explored values of $\rho(\Wh)$ in $\{0.1, 0.2, \ldots, 1.5 \}$,  $\omega_{x}$ and $\omega_{b}$
in $\{10^{j} : j = -3, \ldots, 1\}$, and  $\alpha$ in $\{10^{j} : j = -5, \ldots, 0\}$.
Regarding $\Wh$ initialization, for ESNs we used the fast initialization strategy described in \cite{gallicchio2019fastsr}, while for R-ESN all the non-zero elements in $\Wh$ shared the same weight value, as in \cite{Rodan2010minimum}.
In all the cases, 
the readout was implemented by a dense layer
fed by the last state computed by the reservoir for each time-series, and trained in closed-form by ridge regression\footnote{For all the RC models, the reservoir part was implemented in Keras as a custom RNN layer, while the readout component used Scikit-learn RidgeClassifier with the default regularization strength value of 1.}.
The hyper-parameters were fine-tuned by model selection on the validation set using 
random search with $200$ iterations,
individually for each model. 
After model selection, for each model
we generated $10$ instances (random guesses) of the network with the selected configuration. The $10$ instances were trained on the training set and evaluated on the test set. The reported results have been finally achieved by computing averages and standard deviations over the 10 instances.\\

\begin{figure*}
    \centering
    \includegraphics[width=0.7\textwidth]{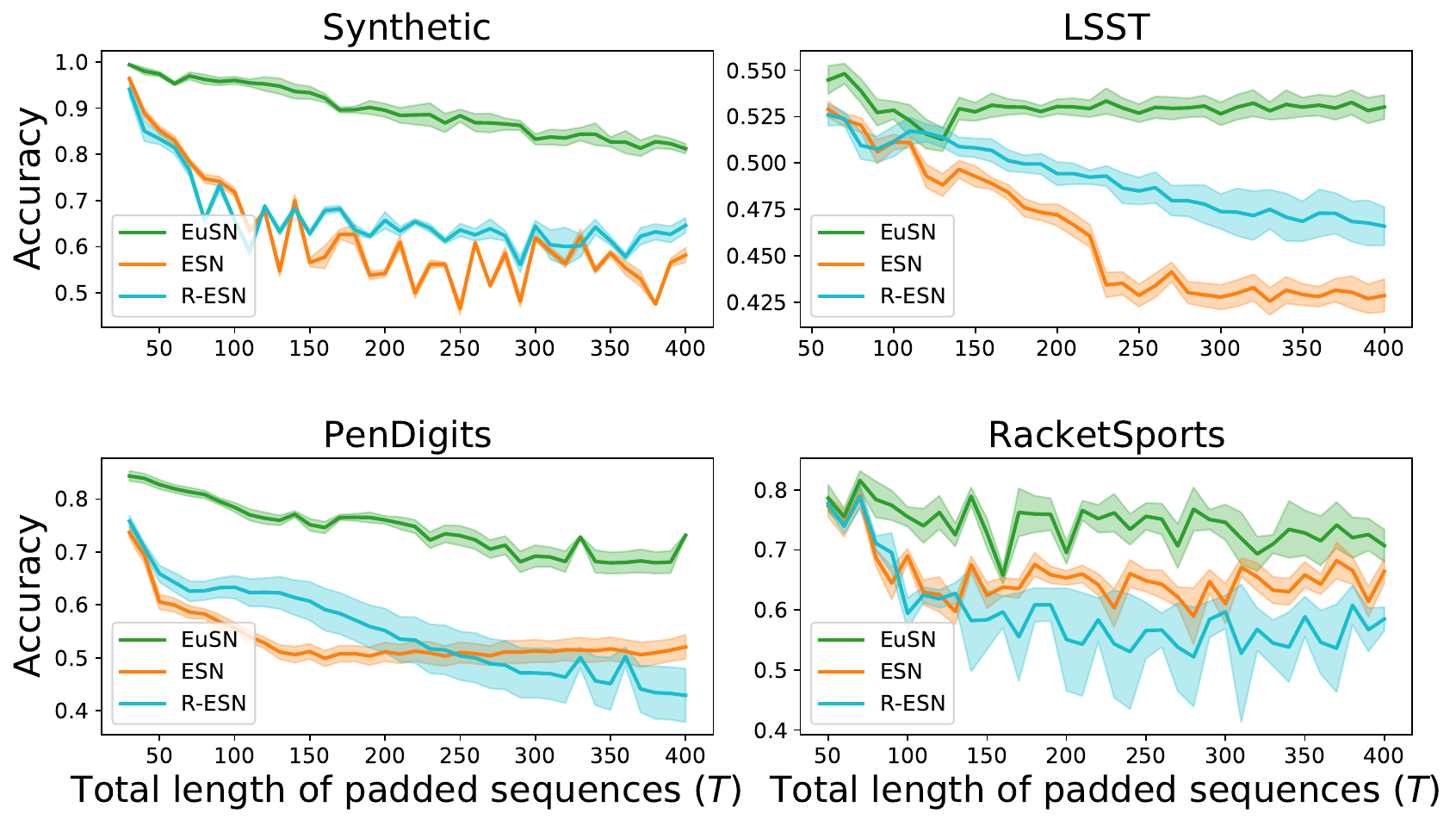}
    \caption{Accuracy achieved on the LTM tasks by EuSN for increasing length of the total length of padded sequences $T$.
    Results are compared to standard ESN and R-ESN. 
    For each value of $T$, the plots report the accuracy values averaged over the 10 random instances, and the corresponding standard deviation as a shaded area.}
    \label{fig.memory_results}
\end{figure*}

\noindent
\textbf{Results -- }
The results on the LTM tasks are reported in Figure~\ref{fig.memory_results}, which shows the test set accuracy achieved by EuSN, ESN and R-ESN, for increasing values of the total length of padded sequences. Each plot in Figure~\ref{fig.memory_results} corresponds to one of the LTM tasks considered in this study.
As it is evident, the performance shown by EuSN is consistently better than that of ESN and R-ESN.
In general, it can be observed that the level of accuracy obtained by all models decreases as $T$ increases. Crucially, the accuracy of EuSN decreases slowly and remains high even for high values of $T$, 
indicating an effective propagation process of task-relevant information even after hundreds of time-steps. 
In contrast, 
consistent with its fading memory characterization,
ESN (and R-ESN) shows a rapidly degrading performance. For example, in the case of the Synthetic task, ESN's accuracy is close to chance level (i.e., $\approx 0.5$) after a few hundred time-steps.

Overall, results in Figure~\ref{fig.memory_results} clearly indicate that EuSN are able to propagate the input information effectively in tasks requiring long-term memorization abilities, 
overcoming the fading memory limitations which are typical of ESN variants.




\subsection{Time-series Classification Benchmarks}
\label{sec.benchmarks}

We assessed the performance of EuSN on several real-world time-series classification benchmarks of diverse nature.
The first 17 datasets were taken from the UEA \& UCR time-series classification repository \cite{timeseriesdatasets}, namely: \emph{Adiac} \cite{jalba2004automatic}, 
\emph{Blink} \cite{chicaiza2021brain},
\emph{CharacterTrajectories} \cite{williams2006extracting}, 
\emph{ECG5000} \cite{goldberger2000physiobank},
\emph{Epilepsy} \cite{villar2016generalized}, 
\emph{FordA},
\emph{HandOutlines} \cite{davis2012segmentation},
\emph{Hearthbeat} \cite{goldberger2000physiobank},
\emph{Libras} \cite{dias2009hand}, 
\emph{Mallat} \cite{mallat1999wavelet},
\emph{MotionSenseHAR} \cite{malekzadeh2019mobile}
\emph{ShapesAll} \cite{latecki2000shape}, 
\emph{SpokenArabicDigits} \cite{hammami2010improved},
\emph{Trace},
\emph{UWaveGestureLibraryAll} \cite{liu2009uwave},
\emph{Wafer} \cite{olszewski2001generalized},
and
\emph{Yoga}.
Moreover, we have considered the 
\emph{IMDB} movie review sentiment classification dataset
\cite{maas-EtAl:2011:ACL-HLT2011}, and the
\emph{Reuters} newswire classification dataset from UCI \cite{apte1994automated}. These two datasets were utilized in the parsed version that is made publicly available online\footnote{IMDB:
\url{https://keras.io/api/datasets/imdb/};
Reuters: \url{https://keras.io/api/datasets/reuters/}}. In addition, individually for the two tasks, we have applied a pre-processing phase to represent each sentence by a time-series of $32$-dimensional word embeddings\footnote{
Each sentence was represented by a sequence of words among the $10000$ most frequent words in the database, with truncation to the maximum length of $200$. Word embeddings have been obtained by training an MLP architecture with preliminary embedding layer with $32$ units, a hidden layer containing $128$ units with ReLU activation, and a final dense output layer (with $1$ sigmoidal unit for IMDB, and $46$ softmax units for Reuters).
This architecture 
was trained on the training set with RMSProp for $100$ epochs, using early stopping with patience $10$ on a validation set containing the $33\%$ of the original training data. The output of the embedding layer on the sentences in the dataset is then used as input feature for our experiments.}.
Finally, we performed experiments with a pixel-by-pixel (i.e., sequential) version of the \emph{MNIST} dataset \cite{lecun1998gradient}, in which each $28 \times 28$ pixels image is reshaped into a $784$-long time-series of gray levels, rescaled to the $0 - 1$ range.

For all datasets, we have used the original data splitting into training and test,
with a further stratified splitting of the original training data into training ($67\%$) and validation ($33\%$) sets.
A summary of the relevant information for each dataset is given in Table~\ref{tab.datasets_benchmarks}.\\

\begin{table*}[t]
\centering
\footnotesize
\begin{tabular}{lrrrrr}
\hline
\textbf{Name}& 
\textbf{\#Seq Tr}&
\textbf{\#Seq Ts}&
\textbf{Length}&
\textbf{Feat.}&
\textbf{Classes}\\
\hline
Adiac & 390 & 391 & 176 & 1 & 37 \\
Blink & 500 & 450 & 510 & 4 & 2 \\
CharacterTrajectories & 1422 & 1436 & 182 & 3 & 20 \\
ECG5000 & 500 & 4500 & 140 & 1 & 5 \\
Epilepsy & 137 & 138 & 206 & 3 & 4 \\
FordA & 3601 & 1320 & 500 & 1 & 2 \\
HandOutlines & 1000 & 370 & 2709 & 1 & 2 \\
Heartbeat & 204 & 205 & 405 & 61 & 2 \\
IMDB & 25000 & 25000 & 200 & 32 & 2 \\
Libras & 180 & 180 & 45 & 2 & 15 \\
Mallat & 55 & 2345 & 1024 & 1 & 8 \\
MNIST & 60000 & 10000 & 784 & 1 & 10 \\
MotionSenseHAR & 966 & 265 & 1000 & 12 & 6 \\
Reuters & 8982 & 2246 & 200 & 32 & 46 \\
ShapesAll & 600 & 600 & 512 & 1 & 60 \\
SpokenArabicDigits & 6599 & 2199 & 93 & 13 & 10 \\
Trace & 100 & 100 & 275 & 1 & 4 \\
UWaveGestureLibraryAll & 896 & 3582 & 945 & 1 & 8 \\
Wafer & 1000 & 6164 & 152 & 1 & 2 \\
Yoga & 300 & 3000 & 426 & 1 & 2 \\
\end{tabular}
\caption{
Information on the time-series classification benchmarks used in the paper, namely:
size of the training (\#Seq Tr) and of the test set (\#Seq Ts), length of the time-series (Length), number of input features, i.e., input dimensionality $X$ (Feat.), and number of class labels (Classes).}
\label{tab.datasets_benchmarks}
\end{table*}

\noindent
\textbf{Experimental Settings --}
We have run experiments with EuSNs on the above mentioned time-series  classification tasks following the same base experimental setting already introduced in Section~\ref{sec.memory} for the LTM tasks.
In this case, however, 
we extended the extent of the comparative analysis to a pool of fully trainable RNN models, including (vanilla) RNNs, A-RNN \cite{chang2019antisymmetricrnn}, 
and Gated Recurrent Unit (GRU) \cite{chung2014empirical}. 
For these models, the output (readout) layer was implemented by a dense layer with 1 sigmoid unit for the binary classification tasks, and a number of softmax units equal to the number of target classes for multi-classification tasks. Analogously, we used binary or categorical cross-entropy as a loss function during training, depending on the number of target classes.
To train these models we used Adam \cite{kingma2014adam} with a maximum number of $5000$ epochs, learning rate chosen from $\{10^{j}:j = -5, \ldots, -1\}$, batch size chosen from $\{10^{j}:j = 5, \ldots, 8\}$ and early stopping with patience $50$.
Similarly to the experimental setup reported Section~\ref{sec.memory}, for these experiments, we performed model selection on the validation set individually for all models. For this, we used random search up to 200 iterations or $10$ hours of computation time.
For each task, after model selection, the performance of each model (in the selected configuration) was assessed through 10 random instances (i.e., guesses), trained on the training set and evaluated on the test set.

In addition to the accuracy results, for every task and every model we measured the
required computational times.
Moreover, to evaluate the energy footprint, we measured
the energy consumed for running all the experiments, integrating the power draw of the GPU (measured in W), sampled every 10 seconds using the NVIDIA System Management Interface.

\begin{figure*}
    \centering
    \includegraphics[width=0.9\textwidth]{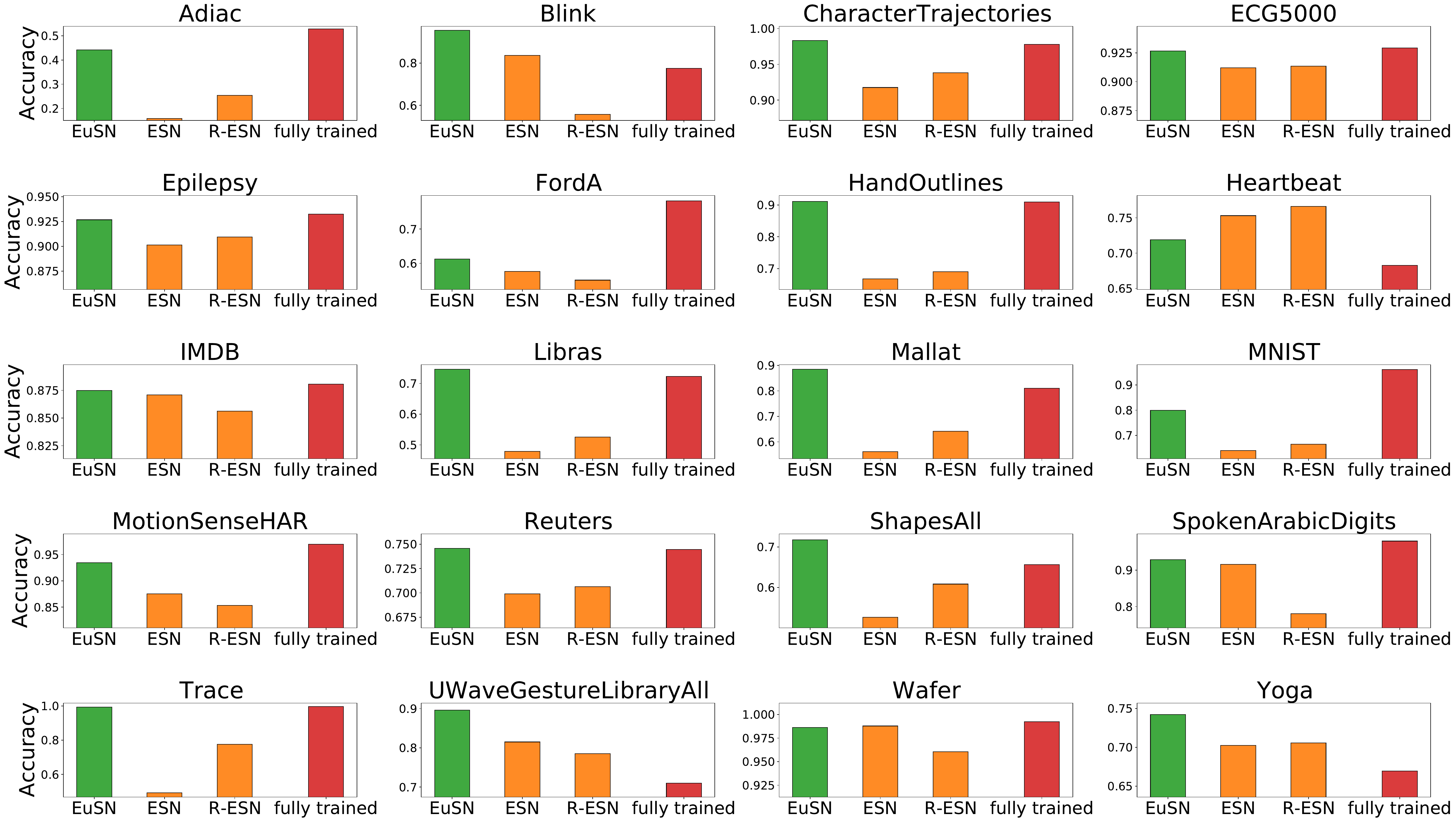}
    \caption{Averaged test set accuracy on the time-series classification benchmarks. 
    The ``fully trained'' results refer to the trainable model (among RNN, A-RNN and GRU) that achieves the highest accuracy on each task. Further details can be found in~\ref{sec.appendix1}.}
    \label{fig.accuracy_results}
\end{figure*}
\begin{figure*}[t!]
    \centering
    \includegraphics[width=0.9\textwidth]{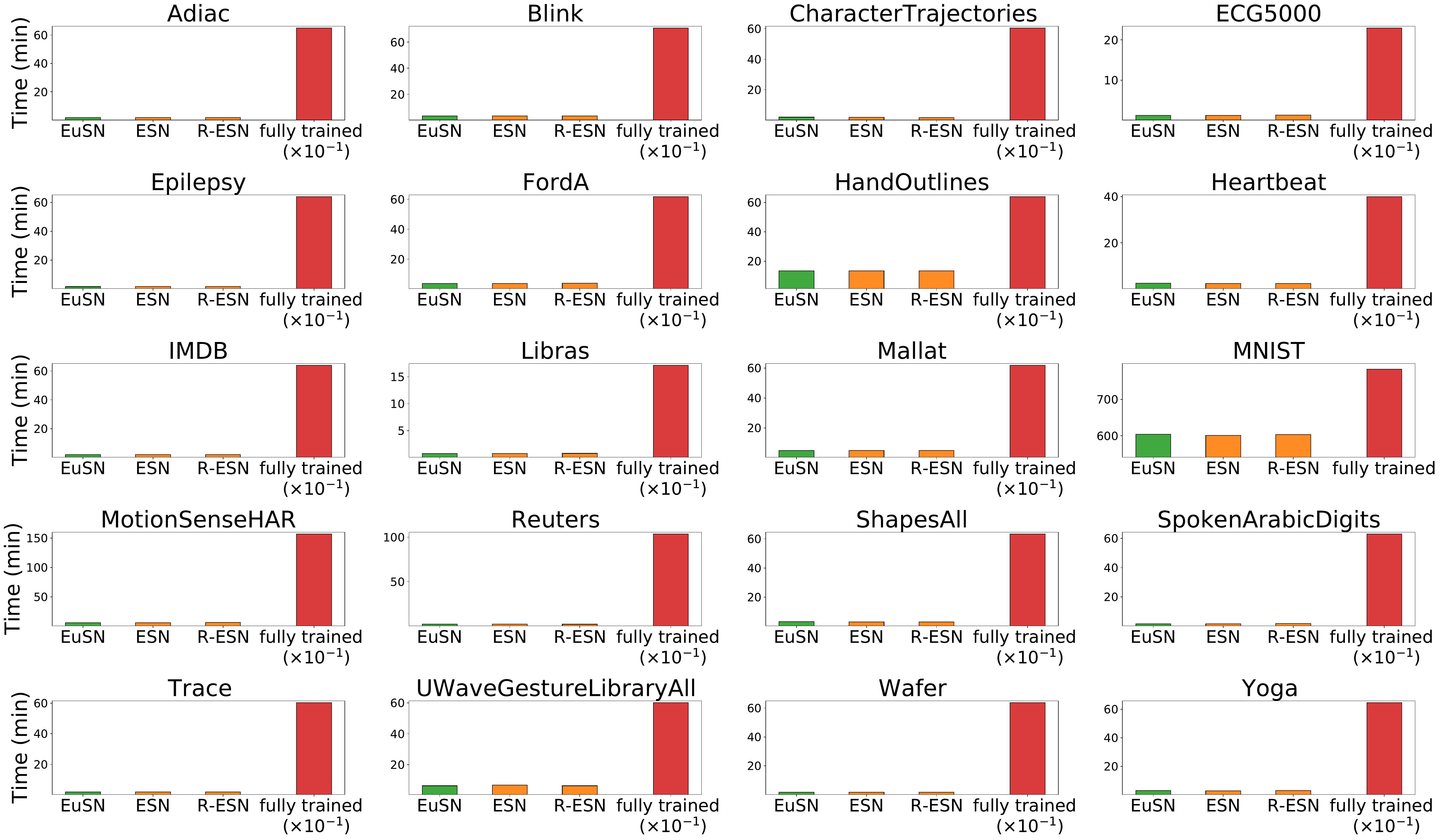}
    \caption{Time required for model selection (in minutes) on the time-series classification benchmarks. The ``fully trained'' results refer to the trainable model (among RNN, A-RNN and GRU) that achieves the highest accuracy on each task. Further details can be found in~\ref{sec.appendix1}.}
    \label{fig.time_results}
\end{figure*}
\begin{figure*}[t!]
    \centering
    \includegraphics[width=0.9\textwidth]{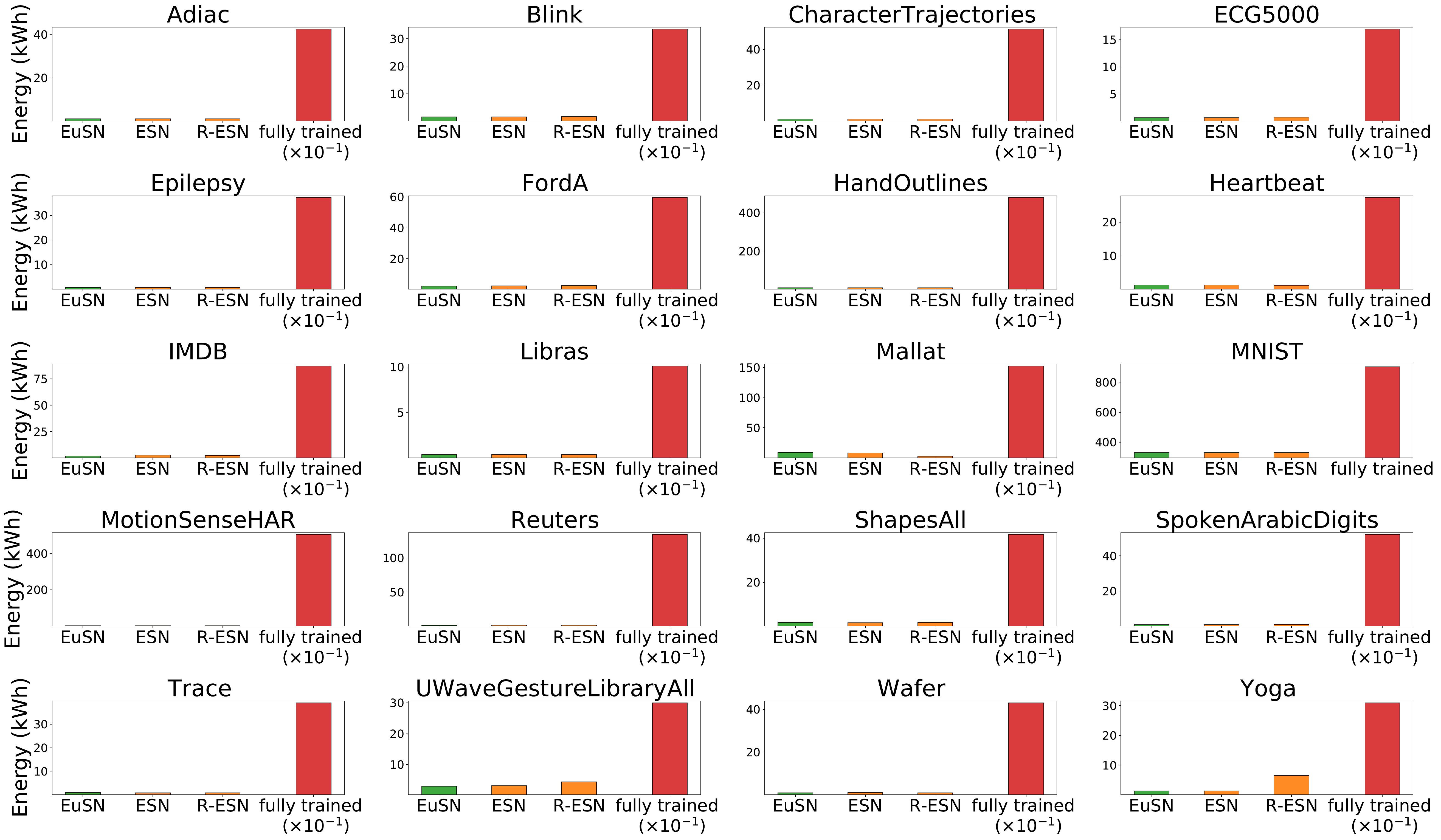}
    \caption{Energy consumption (in kWh) required for the experiments on the time-series classification benchmarks. The ``fully trained'' results refer to the trainable model (among RNN, A-RNN and GRU) that achieves the highest accuracy on each task.
    For ease of visualization, the results of fully trainable models are shown downscaled by a factor of $10$ in all tasks except MNIST.
    Further details can be found in~\ref{sec.appendix1}.}
    \label{fig.energy_results}
\end{figure*}

\noindent
\textbf{Results --} 
For the sake of cleanliness of results presentation, here we report a graphical illustration of the main results of the performed experiments.
The reader can find
full details 
on the obtained results
reported in~\ref{sec.appendix1}.
%
Figure~\ref{fig.accuracy_results} shows the test set accuracy achieved by the considered models on the time-series classification benchmarks. Figures \ref{fig.time_results}, and 
\ref{fig.energy_results}, which follow a similar trend with each other, respectively illustrate the times needed for model selection (in minutes), and the energy consumption required for running the experiments (in kWh). In all three figures, for simplicity's sake, we report a single value for the fully trained models (i.e., RNN, A-RNN, and GRU): the one corresponding to the model that, among them, is found to have the highest accuracy performance. Please refer to~\ref{sec.appendix1} for a detailed report on the results obtained individually for each of the fully trainable models.

Two major observations can be drawn from the results.
First, it is evident from Figure~\ref{fig.accuracy_results} that EuSN in general considerably outperforms both ESN and R-ESN in terms of accuracy.
On the one hand, in tasks where a clear performance gap in favor of fully trained models over standard RC models is appreciable (e.g., Adiac, HandOutlines, Libras, Mallat, MNIST, MotionSenseHAR, Reuters, ShapesAll, Trace), EuSNs allow to largely bridge this gap, in some cases even exceeding the performance of fully trained networks.
On the other hand, in tasks where standard RC models perform better than fully trained ones (Blink, UWaveGestureLibraryAll, Yoga), EuSNs generally allow an even higher performance. The only two exceptions are the Heartbeat and Wafer tasks, where the performance of the standard RC models is better than that of EuSNs. In these cases, however, we can see that EuSNs reach or even exceeds the performance of the fully trained models.
Narrowing our attention to the baseline RC models, we can also observe that the highest accuracy was obtained in most tasks by the R-ESN model, thus confirming the goodness of this architectural variant against standard ESN in applications.

Second,  looking at Figures~\ref{fig.time_results} and \ref{fig.energy_results}, we can see that leveraging untrained dynamics, EuSN proved to be as efficient as standard RC networks, and far more efficient than fully trainable RNNs. Overall, compared to fully trainable models, while showing a competitively high accuracy, EuSN allowed a dramatic advantage in terms of required computational resources, enabling up to an 
$\approx 490$x reduction in computational times,
and an $\approx 1750$x reduction in energy consumption. In contrast, standard RC models, while providing a similar computational advantage, do not achieve the same competitive level of classification performance.





\section{Conclusions}
\label{sec.conclusions}
In this paper we have introduced the Euler State Network (EuSN) model, a novel approach to design RC neural networks, based on the numerical discretization of ODEs by the forward Euler method. The proposed approach leverages the idea of constraining the recurrent reservoir weight matrix to have an antisymmetric structure, resulting in dynamics that are  neither lossy nor unstable,
and allowing for an effective transmission of input signals over time.
Moreover, being featured by untrained recurrent neurons, EuSNs combine the ability to model long-term propagation of input signals with the efficiency typical of RC systems.

Our mathematical analysis of reservoir dynamics revealed that EuSNs are biased towards unitary effective spectral radius and zero local Lyapunov exponents, indicating an intrinsically critical dynamical regime near the edge of stability. 
Experiments on long-term memorization tasks have shown the effectiveness of EuSN in latching the input information into the state representations over long time spans, overcoming the inherent limitations of baseline RC models.
Furthermore, through experiments on time-series classification benchmarks, we found that EuSN provides a formidable trade-off between accuracy and efficiency, compared to state-of-the-art recurrent neural models. In fact, while reaching -- or even outperforming -- the highest level of accuracy achieved by the fully trainable models, the proposed EuSN allows 
a tremendous reduction in computation time and energy consumption. While standard RC approaches can provide similar computational gains, the accuracy levels these achieve are generally considerably lower.

The work presented in this paper demonstrated the validity of an RC-based approach that relies on a profoundly different architectural bias than that of conventional ESNs. In fact, while ESNs represent dynamical systems with fading memory controlled by the spectral properties of the recurrent weight matrix, EuSNs represent dynamical systems with dynamics controllably close to the critical behavior (i.e., to the edge of stability) by construction.
Consequently, while ESNs are naturally able to distinguish input sequences in a suffix-based way, EuSNs are inherently able to distinguish input sequences based on their entire temporal evolution. The investigation carried out in this article showed clear experimental evidence regarding the effectiveness of the proposed method. However, it is important to emphasize that EuSNs are not proposed here as a replacement for ESNs. Instead, they intend to represent a new model in the RC landscape, suitable for problems involving some form of long-term memorization rather than fading memory. For example, in their current form, EuSNs appear unlikely to be suitable for dealing with problems in which it is necessary to model a fast-reaction of the target with respect to changes in the input. Investigations into the Markovian properties of the dynamics of EuSNs represent insights for future work.

Overall, the work presented in this paper is seminal
and 
represents 
a first step in the study of ODE-inspired RC architectures and their theoretical properties. 
Although the presented results are already very encouraging, new directions for future works can be envisaged, for example deepening the study of the differences between ESN and EuSN in terms of reservoir state space organization and nature of the memory structure. 
Other examples of 
future studies may involve an in-depth 
analysis 
of the reservoir topology in EuSN. For example, it might be useful to investigate ways to simplify the architectural construction of the reservoir, while maintaining its computational characteristics, or to expand the diversity of generated dynamics by randomizing the diffusion term among different reservoir neurons. 
Future work would also deserve to explore implementations in neuromorphic hardware, a possibility at once intriguing and challenging, for example in the treatment of noise and the consequences it might have in the realization of non-dissipative 
physical EuSN reservoirs.
Finally, we consider of great interest the 
extensions of the methodology proposed in this paper to graph processing. RC-based approaches have previously shown great potential in this regard, 
resulting in an unparalleled trade-off between application effectiveness and computational efficiency in graph classification \cite{gallicchio2019fast, gallicchio2010graph} and spatio-temporal information processing \cite{cini2022scalable}. In this context, the adoption of an architectural bias based on stable and non-dissipative dynamics could have a decisive impact on tasks that require effective propagation of information through long paths on a graph.
The recent developments described in \cite{gravina2023ICLR}
already show theoretical and empirical evidence of possible benefits, 
and pave the way to further developments in the field of deep graph networks.

\section*{Acknowledgements}
This work is partially supported by the EC H2020 programme under project TEACHING (grant n. 871385).
\bibliographystyle{elsarticle-num} 
\bibliography{references}

\appendix

\section{Detailed Results}
\label{sec.appendix1}

Here, we provide the full details of the experimental outcomes on the time-series classification datasets described in Section~\ref{sec.benchmarks}.

First, in Table~\ref{tab.datasets_ranking} we provide a general overview on the achieved levels of accuracies, by reporting the averaged ranking across the accuracy results on all the time-series classification benchmarks considered.
It can be noticed that EuSNs overall result in the best performing model, followed by A-RNNs and GRUs. Moreover, R-ESNs generally perform better than ESNs, which, along with RNNs are characterized by the worst results in terms of accuracy.
\begin{table}[tbh]
\footnotesize
\centering
\begin{tabular}{lr}
\hline
\emph{Model} & \emph{Avg Rank} \\
\hline
    \vspace{1mm}
EuSN (ours) & 1.80 \\
ESN & 4.15 \\
R-ESN & 3.95 \\
RNN & 4.60 \\
A-RNN & 2.95 \\
GRU & 3.35 \\
\end{tabular}
\caption{Average ranking across all the time-series classification benchmarks.}
\label{tab.datasets_ranking}
\end{table}

In Tables~\ref{tab.datasets_Adiac}-\ref{tab.datasets_Yoga}, we report the full details of the experimental outcomes on the time-series classification datasets described in Section~\ref{sec.benchmarks}. 
For every task and for every model, results in the tables include the following data:
\begin{itemize}
    \item \emph{Acc} - the test set accuracy (averaged over the 10 random guesses, and indicating the standard deviations); 
    \item \emph{\#Par} - the number of trainable parameters of the selected configuration;
    \item \emph{Time (min.)} - the time in minutes required for training and testing the selected configuration (averaged over the 10 random guesses, and indicating the standard deviations);
    \item \emph{MS Time (min.)} - the time in minutes required for model selection;
    \item \emph{Energy (kWH)} - the energy in kWH consumed to perform all the experiments.
\end{itemize}
Notice that, differently from the other models, the GRU implementation could exploit the optimized cuDNN code available through Keras (resulting in a demand of computational resources that is far lower than that which would have resulted from a custom implementation).

Results in Tables~\ref{tab.datasets_Adiac}-\ref{tab.datasets_Yoga}  provide further support to the analysis reported in Section \ref{sec.benchmarks}. Considering also the standard deviation over the results, we can observe that EuSNs substantially outperform both standard ESNs and R-ESNs in terms of predicted accuracy in almost all tasks, at the same time requiring about the same cost in terms of both time and energy consumption. Widening the comparison to fully trained models, EuSNs achieve comparable, and in some cases better, accuracy at costs that are orders of magnitude lower.



\begin{table*}[t]
\centering
\begin{tabular}{lrrrrr}
\hline
\multicolumn{6}{c}{\textbf{Adiac}} \\
    \hline
    & \emph{Acc} & \emph{\# Par} & \emph{Time (min.)} &
      \emph{MS Time (min.)} & \emph{Energy (kWh)}\\
    \vspace{1mm}
EuSN (ours) & $0.442$ ($\pm 0.009$)
	 & 7437 & $0.011$ ($\pm 0.001$)
	 & $1.834$ & $1.028$ \\
ESN & $0.159$ ($\pm 0.067$)
	 & 5957 & $0.010$ ($\pm 0.001$)
	 & $1.921$ & $1.046$ \\
R-ESN & $0.254$ ($\pm 0.007$)
	 & 7067 & $0.010$ ($\pm 0.001$)
	 & $1.888$ & $1.037$ \\
RNN & $0.243$ ($\pm 0.105$)
	 & 43547 & $22.808$ ($\pm 8.195$)
	 & $621.018$ & $484.170$ \\
A-RNN & $0.528$ ($\pm 0.068$)
	 & 35567 & $9.500$ ($\pm 1.193$)
	 & $648.365$ & $424.210$ \\
GRU & $0.201$ ($\pm 0.146$)
	 & 28477 & $0.341$ ($\pm 0.134$)
	 & $479.226$ & $351.676$ \\
\end{tabular}
\caption{Results on the Adiac dataset.}
\label{tab.datasets_Adiac}
\end{table*}
 
\begin{table*}[t]
\centering
\begin{tabular}{lrrrrr}
\hline
\multicolumn{6}{c}{\textbf{Blink}} \\
    \hline
    & \emph{Acc} & \emph{\# Par} & \emph{Time (min.)} &
      \emph{MS Time (min.)} & \emph{Energy (kWh)}\\
    \vspace{1mm}
EuSN (ours) & $0.955$ ($\pm 0.010$)
	 & 131 & $0.019$ ($\pm 0.001$)
	 & $3.530$ & $1.587$ \\
ESN & $0.837$ ($\pm 0.028$)
	 & 171 & $0.021$ ($\pm 0.002$)
	 & $3.539$ & $1.587$ \\
R-ESN & $0.557$ ($\pm 0.033$)
	 & 131 & $0.020$ ($\pm 0.001$)
	 & $3.499$ & $1.666$ \\
RNN & $0.478$ ($\pm 0.044$)
	 & 10601 & $16.352$ ($\pm 8.338$)
	 & $602.812$ & $349.893$ \\
A-RNN & $0.775$ ($\pm 0.044$)
	 & 6881 & $2.788$ ($\pm 0.783$)
	 & $707.131$ & $334.985$ \\
GRU & $0.658$ ($\pm 0.037$)
	 & 53171 & $6.670$ ($\pm 1.485$)
	 & $613.643$ & $312.688$ \\
\end{tabular}
\caption{Results on the Blink dataset.}
\label{tab.datasets_Blink}
\end{table*}
 
\begin{table*}[t]

\centering
\begin{tabular}{lrrrrr}
\hline
\multicolumn{6}{c}{\textbf{CharacterTrajectories}} \\
    \hline
    & \emph{Acc} & \emph{\# Par} & \emph{Time (min.)} &
      \emph{MS Time (min.)} & \emph{Energy (kWh)}\\
    \vspace{1mm}
EuSN (ours) & $0.983$ ($\pm 0.002$)
	 & 3620 & $0.011$ ($\pm 0.001$)
	 & $2.012$ & $1.153$ \\
ESN & $0.918$ ($\pm 0.012$)
	 & 3220 & $0.011$ ($\pm 0.000$)
	 & $2.000$ & $1.069$ \\
R-ESN & $0.938$ ($\pm 0.008$)
	 & 3620 & $0.011$ ($\pm 0.001$)
	 & $1.782$ & $1.093$ \\
RNN & $0.684$ ($\pm 0.152$)
	 & 14760 & $6.087$ ($\pm 2.627$)
	 & $601.495$ & $376.926$ \\
A-RNN & $0.978$ ($\pm 0.003$)
	 & 1640 & $18.828$ ($\pm 1.652$)
	 & $932.676$ & $633.714$ \\
GRU & $0.978$ ($\pm 0.004$)
	 & 92670 & $0.438$ ($\pm 0.153$)
	 & $603.785$ & $512.404$ \\
\end{tabular}
\caption{Results on the CharacterTrajectories dataset.}
\label{tab.datasets_CharacterTrajectories}
\end{table*}
 
\begin{table*}[t]

\centering
\begin{tabular}{lrrrrr}
\hline
\multicolumn{6}{c}{\textbf{ECG5000}} \\
    \hline
    & \emph{Acc} & \emph{\# Par} & \emph{Time (min.)} &
      \emph{MS Time (min.)} & \emph{Energy (kWh)}\\
    \vspace{1mm}
EuSN (ours) & $0.927$ ($\pm 0.002$)
	 & 855 & $0.009$ ($\pm 0.001$)
	 & $1.287$ & $0.666$ \\
ESN & $0.912$ ($\pm 0.001$)
	 & 1005 & $0.009$ ($\pm 0.001$)
	 & $1.266$ & $0.664$ \\
R-ESN & $0.913$ ($\pm 0.002$)
	 & 755 & $0.008$ ($\pm 0.001$)
	 & $1.326$ & $0.750$ \\
RNN & $0.923$ ($\pm 0.011$)
	 & 1885 & $1.557$ ($\pm 0.441$)
	 & $588.373$ & $343.755$ \\
A-RNN & $0.920$ ($\pm 0.006$)
	 & 30095 & $1.992$ ($\pm 0.158$)
	 & $630.257$ & $373.438$ \\
GRU & $0.929$ ($\pm 0.003$)
	 & 60765 & $0.257$ ($\pm 0.020$)
	 & $229.285$ & $169.291$ \\
\end{tabular}
\caption{Results on the ECG5000 dataset.}
\label{tab.datasets_ECG5000}
\end{table*}
 
\begin{table*}[t]

\centering
\begin{tabular}{lrrrrr}
\hline
\multicolumn{6}{c}{\textbf{Epilepsy}} \\
    \hline
    & \emph{Acc} & \emph{\# Par} & \emph{Time (min.)} &
      \emph{MS Time (min.)} & \emph{Energy (kWh)}\\
    \vspace{1mm}
EuSN (ours) & $0.927$ ($\pm 0.012$)
	 & 684 & $0.008$ ($\pm 0.001$)
	 & $1.527$ & $0.752$ \\
ESN & $0.901$ ($\pm 0.017$)
	 & 684 & $0.009$ ($\pm 0.001$)
	 & $1.557$ & $0.842$ \\
R-ESN & $0.909$ ($\pm 0.009$)
	 & 764 & $0.008$ ($\pm 0.001$)
	 & $1.585$ & $0.840$ \\
RNN & $0.560$ ($\pm 0.121$)
	 & 23704 & $1.500$ ($\pm 0.541$)
	 & $215.733$ & $130.642$ \\
A-RNN & $0.933$ ($\pm 0.008$)
	 & 7044 & $1.340$ ($\pm 0.255$)
	 & $639.625$ & $371.542$ \\
GRU & $0.704$ ($\pm 0.152$)
	 & 61464 & $0.171$ ($\pm 0.034$)
	 & $181.629$ & $121.691$ \\
\end{tabular}
\caption{Results on the Epilepsy dataset.}
\label{tab.datasets_Epilepsy}
\end{table*}
 
\begin{table*}[t]

\centering
\begin{tabular}{lrrrrr}
\hline
\multicolumn{6}{c}{\textbf{FordA}} \\
    \hline
    & \emph{Acc} & \emph{\# Par} & \emph{Time (min.)} &
      \emph{MS Time (min.)} & \emph{Energy (kWh)}\\
    \vspace{1mm}
EuSN (ours) & $0.612$ ($\pm 0.015$)
	 & 191 & $0.020$ ($\pm 0.002$)
	 & $3.766$ & $2.350$ \\
ESN & $0.576$ ($\pm 0.012$)
	 & 201 & $0.022$ ($\pm 0.001$)
	 & $3.843$ & $2.515$ \\
R-ESN & $0.551$ ($\pm 0.018$)
	 & 151 & $0.022$ ($\pm 0.002$)
	 & $3.977$ & $2.554$ \\
RNN & $0.782$ ($\pm 0.140$)
	 & 36671 & $37.616$ ($\pm 31.107$)
	 & $616.725$ & $596.350$ \\
A-RNN & $0.494$ ($\pm 0.013$)
	 & 6641 & $29.183$ ($\pm 13.680$)
	 & $1780.866$ & $1252.646$ \\
GRU & $0.717$ ($\pm 0.184$)
	 & 52001 & $16.305$ ($\pm 9.991$)
	 & $447.561$ & $512.732$ \\
\end{tabular}
\caption{Results on the FordA dataset.}
\label{tab.datasets_FordA}
\end{table*}
 
\begin{table*}[t]

\centering
\begin{tabular}{lrrrrr}
\hline
\multicolumn{6}{c}{\textbf{HandOutlines}} \\
    \hline
    & \emph{Acc} & \emph{\# Par} & \emph{Time (min.)} &
      \emph{MS Time (min.)} & \emph{Energy (kWh)}\\
    \vspace{1mm}
EuSN (ours) & $0.911$ ($\pm 0.004$)
	 & 191 & $0.067$ ($\pm 0.003$)
	 & $13.438$ & $8.187$ \\
ESN & $0.668$ ($\pm 0.041$)
	 & 71 & $0.071$ ($\pm 0.003$)
	 & $13.378$ & $8.136$ \\
R-ESN & $0.690$ ($\pm 0.047$)
	 & 61 & $0.069$ ($\pm 0.005$)
	 & $13.418$ & $8.180$ \\
RNN & $0.673$ ($\pm 0.024$)
	 & 2651 & $160.575$ ($\pm 85.782$)
	 & $792.408$ & $1362.865$ \\
A-RNN & $0.909$ ($\pm 0.005$)
	 & 32941 & $706.254$ ($\pm 305.364$)
	 & $637.913$ & $4789.139$ \\
GRU & $0.659$ ($\pm 0.015$)
	 & 401 & $5.151$ ($\pm 1.338$)
	 & $845.320$ & $806.848$ \\
\end{tabular}
\caption{Results on the HandOutlines dataset.}
\label{tab.datasets_HandOutlines}
\end{table*}
 
\begin{table*}[t]

\centering
\begin{tabular}{lrrrrr}
\hline
\multicolumn{6}{c}{\textbf{Heartbeat}} \\
    \hline
    & \emph{Acc} & \emph{\# Par} & \emph{Time (min.)} &
      \emph{MS Time (min.)} & \emph{Energy (kWh)}\\
    \vspace{1mm}
EuSN (ours) & $0.719$ ($\pm 0.021$)
	 & 171 & $0.013$ ($\pm 0.000$)
	 & $2.544$ & $1.414$ \\
ESN & $0.753$ ($\pm 0.008$)
	 & 121 & $0.013$ ($\pm 0.001$)
	 & $2.490$ & $1.425$ \\
R-ESN & $0.766$ ($\pm 0.010$)
	 & 191 & $0.013$ ($\pm 0.001$)
	 & $2.460$ & $1.322$ \\
RNN & $0.670$ ($\pm 0.026$)
	 & 48071 & $9.610$ ($\pm 2.240$)
	 & $614.315$ & $399.403$ \\
A-RNN & $0.633$ ($\pm 0.054$)
	 & 16301 & $11.321$ ($\pm 4.801$)
	 & $619.311$ & $413.130$ \\
GRU & $0.682$ ($\pm 0.040$)
	 & 96001 & $0.419$ ($\pm 0.031$)
	 & $399.335$ & $272.900$ \\
\end{tabular}
\caption{Results on the Heartbeat dataset.}
\label{tab.datasets_Heartbeat}
\end{table*}
 
\begin{table*}[t]

\centering
\begin{tabular}{lrrrrr}
\hline
\multicolumn{6}{c}{\textbf{IMDB}} \\
    \hline
    & \emph{Acc} & \emph{\# Par} & \emph{Time (min.)} &
      \emph{MS Time (min.)} & \emph{Energy (kWh)}\\
    \vspace{1mm}
EuSN (ours) & $0.875$ ($\pm 0.002$)
	 & 201 & $0.016$ ($\pm 0.001$)
	 & $2.087$ & $2.001$ \\
ESN & $0.871$ ($\pm 0.007$)
	 & 111 & $0.010$ ($\pm 0.000$)
	 & $2.127$ & $2.637$ \\
R-ESN & $0.856$ ($\pm 0.020$)
	 & 111 & $0.011$ ($\pm 0.001$)
	 & $2.092$ & $2.317$ \\
RNN & $0.872$ ($\pm 0.006$)
	 & 2961 & $207.990$ ($\pm 130.754$)
	 & $761.358$ & $1613.372$ \\
A-RNN & $0.863$ ($\pm 0.000$)
	 & 11161 & $272.227$ ($\pm 92.466$)
	 & $783.339$ & $1973.787$ \\
GRU & $0.880$ ($\pm 0.000$)
	 & 93281 & $38.173$ ($\pm 5.213$)
	 & $603.641$ & $872.405$ \\
\end{tabular}
\caption{Results on the IMDB dataset.}
\label{tab.datasets_IMDB}
\end{table*}
 
\begin{table*}[t]

\centering
\begin{tabular}{lrrrrr}
\hline
\multicolumn{6}{c}{\textbf{Libras}} \\
    \hline
    & \emph{Acc} & \emph{\# Par} & \emph{Time (min.)} &
      \emph{MS Time (min.)} & \emph{Energy (kWh)}\\
    \vspace{1mm}
EuSN (ours) & $0.746$ ($\pm 0.031$)
	 & 2565 & $0.006$ ($\pm 0.001$)
	 & $0.768$ & $0.392$ \\
ESN & $0.479$ ($\pm 0.017$)
	 & 2415 & $0.006$ ($\pm 0.001$)
	 & $0.752$ & $0.384$ \\
R-ESN & $0.526$ ($\pm 0.013$)
	 & 2715 & $0.006$ ($\pm 0.001$)
	 & $0.794$ & $0.378$ \\
RNN & $0.597$ ($\pm 0.039$)
	 & 22135 & $0.406$ ($\pm 0.106$)
	 & $341.168$ & $195.617$ \\
A-RNN & $0.721$ ($\pm 0.017$)
	 & 39535 & $0.500$ ($\pm 0.036$)
	 & $600.094$ & $344.389$ \\
GRU & $0.723$ ($\pm 0.175$)
	 & 113445 & $0.132$ ($\pm 0.015$)
	 & $170.894$ & $100.965$ \\
\end{tabular}
\caption{Results on the Libras dataset.}
\label{tab.datasets_Libras}
\end{table*}
 
\begin{table*}[t]

\centering
\begin{tabular}{lrrrrr}
\hline
\multicolumn{6}{c}{\textbf{Mallat}} \\
    \hline
    & \emph{Acc} & \emph{\# Par} & \emph{Time (min.)} &
      \emph{MS Time (min.)} & \emph{Energy (kWh)}\\
    \vspace{1mm}
EuSN (ours) & $0.885$ ($\pm 0.014$)
	 & 1528 & $0.029$ ($\pm 0.001$)
	 & $4.930$ & $9.306$ \\
ESN & $0.562$ ($\pm 0.020$)
	 & 1368 & $0.028$ ($\pm 0.002$)
	 & $4.959$ & $8.097$ \\
R-ESN & $0.642$ ($\pm 0.054$)
	 & 1448 & $0.026$ ($\pm 0.001$)
	 & $5.074$ & $3.029$ \\
RNN & $0.203$ ($\pm 0.079$)
	 & 27208 & $9.314$ ($\pm 2.879$)
	 & $627.158$ & $656.499$ \\
A-RNN & $0.810$ ($\pm 0.017$)
	 & 11008 & $4.631$ ($\pm 0.716$)
	 & $617.484$ & $1525.189$ \\
GRU & $0.179$ ($\pm 0.032$)
	 & 70058 & $0.430$ ($\pm 0.046$)
	 & $183.117$ & $656.499$ \\
\end{tabular}
\caption{Results on the Mallat dataset.}
\label{tab.datasets_Mallat}
\end{table*}
 
\begin{table*}[t]

\centering
\begin{tabular}{lrrrrr}
\hline
\multicolumn{6}{c}{\textbf{MNIST}} \\
    \hline
    & \emph{Acc} & \emph{\# Par} & \emph{Time (min.)} &
      \emph{MS Time (min.)} & \emph{Energy (kWh)}\\
    \vspace{1mm}
EuSN (ours) & $0.800$ ($\pm 0.004$)
	 & 1910 & $3.498$ ($\pm 0.038$)
	 & $603.901$ & $330.898$ \\
ESN & $0.640$ ($\pm 0.009$)
	 & 1110 & $3.676$ ($\pm 0.138$)
	 & $600.850$ & $329.289$ \\
R-ESN & $0.665$ ($\pm 0.009$)
	 & 810 & $3.643$ ($\pm 0.161$)
	 & $603.340$ & $329.365$ \\
RNN & $0.885$ ($\pm 0.153$)
	 & 21290 & $5485.678$ ($\pm 1714.600$)
	 & $7529.925$ & $22767.684$ \\
A-RNN & $0.867$ ($\pm 0.008$)
	 & 7370 & $2178.702$ ($\pm 463.826$)
	 & $2486.857$ & $15664.276$ \\
GRU & $0.961$ ($\pm 0.049$)
	 & 8460 & $79.697$ ($\pm 17.154$)
	 & $782.909$ & $902.842$ \\
\end{tabular}
\caption{Results on the MNIST dataset.}
\label{tab.datasets_MNIST}
\end{table*}
 
\begin{table*}[t]

\centering
\begin{tabular}{lrrrrr}
\hline
\multicolumn{6}{c}{\textbf{MotionSenseHAR}} \\
    \hline
    & \emph{Acc} & \emph{\# Par} & \emph{Time (min.)} &
      \emph{MS Time (min.)} & \emph{Energy (kWh)}\\
    \vspace{1mm}
EuSN (ours) & $0.934$ ($\pm 0.007$)
	 & 1026 & $0.034$ ($\pm 0.002$)
	 & $6.153$ & $2.884$ \\
ESN & $0.875$ ($\pm 0.021$)
	 & 1026 & $0.035$ ($\pm 0.005$)
	 & $5.956$ & $2.809$ \\
R-ESN & $0.853$ ($\pm 0.028$)
	 & 1146 & $0.034$ ($\pm 0.003$)
	 & $6.339$ & $2.961$ \\
RNN & $0.700$ ($\pm 0.102$)
	 & 16686 & $14.150$ ($\pm 7.962$)
	 & $610.699$ & $341.362$ \\
A-RNN & $0.970$ ($\pm 0.006$)
	 & 19376 & $71.593$ ($\pm 6.250$)
	 & $1569.631$ & $5047.365$ \\
GRU & $0.935$ ($\pm 0.035$)
	 & 129606 & $0.508$ ($\pm 0.066$)
	 & $614.305$ & $1527.228$ \\
\end{tabular}
\caption{Results on the MotionSenseHAR dataset.}
\label{tab.datasets_MotionSenseHAR}
\end{table*}
 
\begin{table*}[t]

\centering
\begin{tabular}{lrrrrr}
\hline
\multicolumn{6}{c}{\textbf{Reuters}} \\
    \hline
    & \emph{Acc} & \emph{\# Par} & \emph{Time (min.)} &
      \emph{MS Time (min.)} & \emph{Energy (kWh)}\\
    \vspace{1mm}
EuSN (ours) & $0.746$ ($\pm 0.002$)
	 & 8326 & $0.011$ ($\pm 0.001$)
	 & $2.111$ & $1.226$ \\
ESN & $0.699$ ($\pm 0.003$)
	 & 7866 & $0.009$ ($\pm 0.000$)
	 & $2.215$ & $1.624$ \\
R-ESN & $0.706$ ($\pm 0.003$)
	 & 8786 & $0.012$ ($\pm 0.001$)
	 & $2.086$ & $1.502$ \\
RNN & $0.486$ ($\pm 0.095$)
	 & 4806 & $324.440$ ($\pm 230.949$)
	 & $710.351$ & $2242.565$ \\
A-RNN & $0.745$ ($\pm 0.003$)
	 & 27216 & $119.179$ ($\pm 2.328$)
	 & $1035.142$ & $1343.872$ \\
GRU & $0.724$ ($\pm 0.047$)
	 & 37666 & $2.759$ ($\pm 0.016$)
	 & $630.515$ & $583.792$ \\
\end{tabular}
\caption{Results on the Reuters dataset.}
\label{tab.datasets_Reuters}
\end{table*}
 
\begin{table*}[t]

\centering
\begin{tabular}{lrrrrr}
\hline
\multicolumn{6}{c}{\textbf{ShapesAll}} \\
    \hline
    & \emph{Acc} & \emph{\# Par} & \emph{Time (min.)} &
      \emph{MS Time (min.)} & \emph{Energy (kWh)}\\
    \vspace{1mm}
EuSN (ours) & $0.718$ ($\pm 0.007$)
	 & 10860 & $0.016$ ($\pm 0.001$)
	 & $3.257$ & $1.900$ \\
ESN & $0.526$ ($\pm 0.015$)
	 & 10260 & $0.015$ ($\pm 0.001$)
	 & $3.083$ & $1.724$ \\
R-ESN & $0.608$ ($\pm 0.020$)
	 & 10860 & $0.019$ ($\pm 0.002$)
	 & $3.107$ & $1.825$ \\
RNN & $0.182$ ($\pm 0.053$)
	 & 18980 & $7.641$ ($\pm 2.683$)
	 & $600.022$ & $386.331$ \\
A-RNN & $0.656$ ($\pm 0.018$)
	 & 2820 & $9.672$ ($\pm 0.709$)
	 & $632.344$ & $416.894$ \\
GRU & $0.588$ ($\pm 0.028$)
	 & 59730 & $1.072$ ($\pm 0.131$)
	 & $622.661$ & $540.201$ \\
\end{tabular}
\caption{Results on the ShapesAll dataset.}
\label{tab.datasets_ShapesAll}
\end{table*}
 
\begin{table*}[t]

\centering
\begin{tabular}{lrrrrr}
\hline
\multicolumn{6}{c}{\textbf{SpokenArabicDigits}} \\
    \hline
    & \emph{Acc} & \emph{\# Par} & \emph{Time (min.)} &
      \emph{MS Time (min.)} & \emph{Energy (kWh)}\\
    \vspace{1mm}
EuSN (ours) & $0.929$ ($\pm 0.009$)
	 & 1910 & $0.010$ ($\pm 0.001$)
	 & $1.605$ & $0.927$ \\
ESN & $0.916$ ($\pm 0.009$)
	 & 710 & $0.008$ ($\pm 0.001$)
	 & $1.634$ & $0.964$ \\
R-ESN & $0.781$ ($\pm 0.085$)
	 & 1810 & $0.011$ ($\pm 0.002$)
	 & $1.703$ & $1.074$ \\
RNN & $0.875$ ($\pm 0.053$)
	 & 40670 & $3.906$ ($\pm 0.587$)
	 & $610.520$ & $373.028$ \\
A-RNN & $0.845$ ($\pm 0.003$)
	 & 12410 & $475.437$ ($\pm 11.460$)
	 & $700.678$ & $3117.685$ \\
GRU & $0.979$ ($\pm 0.003$)
	 & 85610 & $0.392$ ($\pm 0.112$)
	 & $629.330$ & $521.702$ \\
\end{tabular}
\caption{Results on the SpokenArabicDigits dataset.}
\label{tab.datasets_SpokenArabicDigits}
\end{table*}
 
\begin{table*}[t]

\centering
\begin{tabular}{lrrrrr}
\hline
\multicolumn{6}{c}{\textbf{Trace}} \\
    \hline
    & \emph{Acc} & \emph{\# Par} & \emph{Time (min.)} &
      \emph{MS Time (min.)} & \emph{Energy (kWh)}\\
    \vspace{1mm}
EuSN (ours) & $0.994$ ($\pm 0.008$)
	 & 764 & $0.009$ ($\pm 0.000$)
	 & $1.969$ & $0.911$ \\
ESN & $0.492$ ($\pm 0.097$)
	 & 484 & $0.009$ ($\pm 0.001$)
	 & $1.902$ & $0.759$ \\
R-ESN & $0.776$ ($\pm 0.052$)
	 & 724 & $0.009$ ($\pm 0.000$)
	 & $1.863$ & $0.835$ \\
RNN & $0.427$ ($\pm 0.161$)
	 & 20444 & $2.878$ ($\pm 1.284$)
	 & $257.409$ & $130.565$ \\
A-RNN & $0.997$ ($\pm 0.005$)
	 & 6884 & $25.065$ ($\pm 0.466$)
	 & $603.131$ & $390.523$ \\
GRU & $0.515$ ($\pm 0.301$)
	 & 11584 & $1.041$ ($\pm 1.815$)
	 & $182.036$ & $87.252$ \\
\end{tabular}
\caption{Results on the Trace dataset.}
\label{tab.datasets_Trace}
\end{table*}
 
\begin{table*}[t]

\centering
\begin{tabular}{lrrrrr}
\hline
\multicolumn{6}{c}{\textbf{UWaveGestureLibraryAll}} \\
    \hline
    & \emph{Acc} & \emph{\# Par} & \emph{Time (min.)} &
      \emph{MS Time (min.)} & \emph{Energy (kWh)}\\
    \vspace{1mm}
EuSN (ours) & $0.896$ ($\pm 0.002$)
	 & 1448 & $0.034$ ($\pm 0.003$)
	 & $6.336$ & $2.961$ \\
ESN & $0.815$ ($\pm 0.016$)
	 & 1368 & $0.033$ ($\pm 0.002$)
	 & $6.751$ & $3.188$ \\
R-ESN & $0.785$ ($\pm 0.006$)
	 & 1128 & $0.033$ ($\pm 0.002$)
	 & $6.331$ & $4.408$ \\
RNN & $0.379$ ($\pm 0.075$)
	 & 9008 & $25.029$ ($\pm 9.022$)
	 & $623.929$ & $399.263$ \\
A-RNN & $0.488$ ($\pm 0.008$)
	 & 208 & $211.754$ ($\pm 4.353$)
	 & $766.737$ & $1313.035$ \\
GRU & $0.710$ ($\pm 0.306$)
	 & 111538 & $5.681$ ($\pm 4.757$)
	 & $601.381$ & $299.809$ \\
\end{tabular}
\caption{Results on the UWaveGestureLibraryAll dataset.}
\label{tab.datasets_UWaveGestureLibraryAll}
\end{table*}
 
\begin{table*}[t]

\centering
\begin{tabular}{lrrrrr}
\hline
\multicolumn{6}{c}{\textbf{Wafer}} \\
    \hline
    & \emph{Acc} & \emph{\# Par} & \emph{Time (min.)} &
      \emph{MS Time (min.)} & \emph{Energy (kWh)}\\
    \vspace{1mm}
EuSN (ours) & $0.986$ ($\pm 0.002$)
	 & 101 & $0.009$ ($\pm 0.001$)
	 & $1.705$ & $0.870$ \\
ESN & $0.988$ ($\pm 0.002$)
	 & 201 & $0.011$ ($\pm 0.000$)
	 & $1.756$ & $1.048$ \\
R-ESN & $0.960$ ($\pm 0.009$)
	 & 61 & $0.009$ ($\pm 0.001$)
	 & $1.759$ & $0.954$ \\
RNN & $0.965$ ($\pm 0.032$)
	 & 26081 & $9.288$ ($\pm 4.885$)
	 & $601.445$ & $397.676$ \\
A-RNN & $0.992$ ($\pm 0.001$)
	 & 10301 & $11.337$ ($\pm 2.300$)
	 & $636.482$ & $430.376$ \\
GRU & $0.981$ ($\pm 0.012$)
	 & 1401 & $0.739$ ($\pm 0.339$)
	 & $318.245$ & $267.189$ \\
\end{tabular}
\caption{Results on the Wafer dataset.}
\label{tab.datasets_Wafer}
\end{table*}
 
\begin{table*}[t]
\centering
\begin{tabular}{lrrrrr}
\hline
\multicolumn{6}{c}{\textbf{Yoga}} \\
    \hline
    & \emph{Acc} & \emph{\# Par} & \emph{Time (min.)} &
      \emph{MS Time (min.)} & \emph{Energy (kWh)}\\
    \vspace{1mm}
EuSN (ours) & $0.742$ ($\pm 0.010$)
	 & 171 & $0.017$ ($\pm 0.001$)
	 & $3.126$ & $1.367$ \\
ESN & $0.702$ ($\pm 0.011$)
	 & 161 & $0.019$ ($\pm 0.001$)
	 & $2.973$ & $1.367$ \\
R-ESN & $0.706$ ($\pm 0.011$)
	 & 151 & $0.018$ ($\pm 0.001$)
	 & $3.099$ & $6.512$ \\
RNN & $0.610$ ($\pm 0.035$)
	 & 3781 & $3.236$ ($\pm 1.095$)
	 & $486.777$ & $1381.872$ \\
A-RNN & $0.669$ ($\pm 0.003$)
	 & 2651 & $3.032$ ($\pm 0.231$)
	 & $645.536$ & $309.413$ \\
GRU & $0.603$ ($\pm 0.051$)
	 & 11401 & $0.334$ ($\pm 0.124$)
	 & $157.973$ & $73.558$ \\
\end{tabular}
\caption{Results on the Yoga dataset.}
\label{tab.datasets_Yoga}
\end{table*}

\end{document}